\documentclass[final]{elsarticle}

\usepackage[edges]{forest}

\usepackage{tabularx}
\usepackage{lineno,hyperref}
\usepackage{color}
\modulolinenumbers[5]

\usepackage{multirow}
\usepackage{graphicx}
\usepackage{subfigure} 
\usepackage{xcolor}
\definecolor{myblue}{RGB}{189,215,238}  
\usepackage[utf8]{inputenc}
\usepackage[TS1,T1]{fontenc}
\usepackage{array, booktabs}
\usepackage{colortbl}
\usepackage{caption}
\definecolor{baseD}{HTML}{7CAFC2}
\newcommand{\foo}{\color{baseD}\makebox[0pt]{\textbullet}\hskip-0.5pt\vrule width 1pt\hspace{\labelsep}}
\usepackage{makecell}
\usepackage{longtable}
\usepackage{booktabs}
\usepackage{textcomp}
\usepackage{amssymb}% http://ctan.org/pkg/amssymb
\usepackage{pifont}% http://ctan.org/pkg/pifont
\newcommand{\cmark}{\ding{51}}%
\newcommand{\xmark}{-}%
\usepackage{awesomebox} % for infobox
\usepackage{bbding}
\usepackage[most]{tcolorbox}

\usepackage[tikz]{bclogo}
\usepackage[framemethod=tikz]{mdframed}
\usepackage{hyperref}
\hypersetup{
    linkcolor=blue,
}

\definecolor{bgblue}{RGB}{245,243,253}
\definecolor{ttblue}{RGB}{91,194,224}

\journal{Journal of \LaTeX\ Templates}

\begin{document}

\begin{frontmatter}

\title{Data Augmentation Approaches in Natural Language Processing: A Survey}


%\iffalse 
% or include affiliations in footnotes:
\author{Bohan Li}
\ead{bhli@ir.hit.edu.cn}

\author{Yutai Hou}
\ead{ythou@ir.hit.edu.cn}

\author{Wanxiang Che\corref{mycorrespondingauthor}} %\thanks{* Corresponding author.}
\cortext[mycorrespondingauthor]{Corresponding author}
\ead{car@ir.hit.edu.cn}

\address{Harbin Institute of Technology, Harbin, China}
%\fi

% ============= yutai version =============
\begin{abstract}
As an effective strategy, data augmentation (DA) alleviates data scarcity scenarios where deep learning techniques may fail. It is widely applied in computer vision then introduced to natural language processing and achieves improvements in many tasks.
One of the main focuses of the DA methods is to improve the diversity of training data, thereby helping the model to better generalize to unseen testing data.
In this survey, we frame DA methods into three categories based on the \textbf{diversity} of augmented data, including paraphrasing, noising, and sampling.
% Unlike direct listing of different methods, such categories inspects the nature of data augmentation.
Our paper sets out to analyze DA methods in detail according to the above categories.
Further, we also introduce their applications in NLP tasks as well as the challenges. Some useful resources are provided in \ref{appendix}.

%As an effective strategy, data augmentation (DA) alleviates data scarcity scenarios where deep learning techniques may fail. It is widely applied in computer vision then introduced to natural language processing and achieves improvements in many tasks. One of the main focuses of the DA methods is to improve the diversity of training data, thereby helping the model to better generalize to unseen testing data. In this survey, we frame DA methods into three categories based on the diversity of augmented data, including paraphrasing, noising, and sampling. Our paper sets out to analyze DA methods in detail according to the above categories. Further, we also introduce their applications in NLP tasks as well as the challenges.
\end{abstract}

% ============ version 1.0 =========
% \begin{abstract}
% As an effective strategy, data augmentation (DA) alleviates data scarcity scenarios where deep learning techniques may fail. It is widely applied in computer vision then introduced to natural language processing and achieve improvements in many tasks. In this survey, we frame DA methods into three categories based on augmented data \textbf{diversity}, including paraphrasing, noising, and sampling. Our paper sets out to analyze DA methods in detail according to the above categories. In addition, we also introduce their applications in NLP tasks as well as the challenges. 
% \end{abstract}


\end{frontmatter}

%\linenumbers

\begin{figure}[h]
    \centering
    \includegraphics[width=0.8\textwidth]{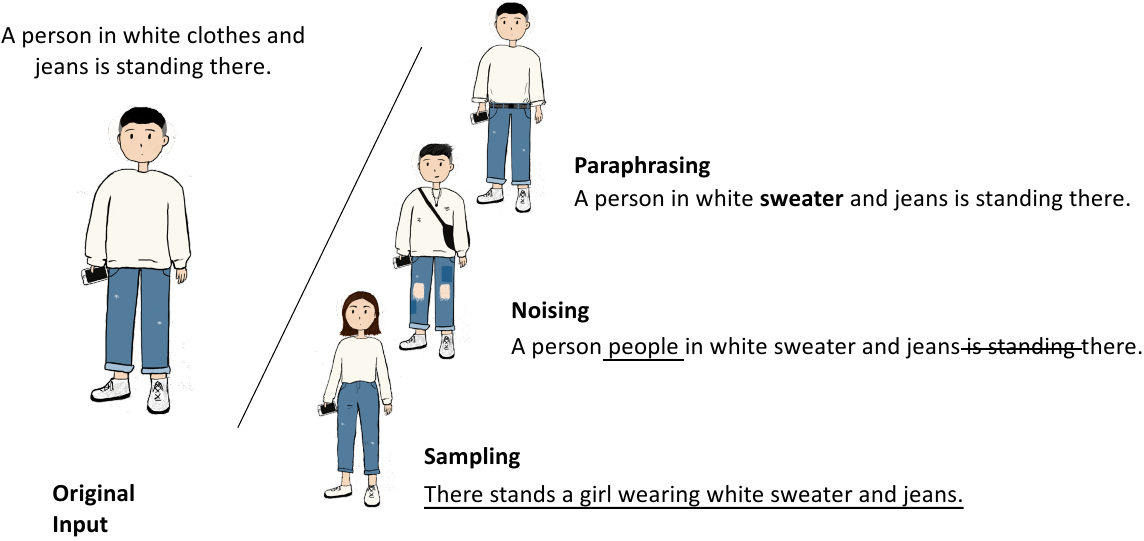}
    %\caption{data augmentation techniques include three categories.}
    \label{Fig.begin}
\end{figure}

%\tableofcontents
%\newpage

\clearpage
\tableofcontents

\clearpage

\section{Introduction}
\label{introduction}
Data augmentation refers to methods used to increase the amount of data by adding slightly modified copies of already existing data or newly created synthetic data from existing data. Such methods alleviate data scarcity scenarios where deep learning techniques may fail, so DA has received active interest and demand recently. Data augmentation is widely applied in the field of computer vision~\cite{shorten2019survey}, such as flipping and rotation, then introduced to natural language processing (NLP). Different to images, natural language is discrete, which makes the adoption of DA methods more difficult and underexplored in NLP.

Large numbers of DA methods have been proposed recently, and a survey of existing methods is beneficial so that researchers could keep up with the speed of innovation. Liu et al.~\cite{Liu2020ASO} and Feng et al.~\cite{feng-etal-2021-survey} both present surveys that give a bird's eye view of DA for NLP. They directly divide the categories according to the methods. These categories thus tend to be too limited or general, e.g., \textit{back-translation} and \textit{model-based techniques}. Bayer et al.~\cite{Bayer2021ASO} post a survey on DA for text classification only. In this survey, we will provide an inclusive overview of DA methods in NLP. One of our main goals is to show the nature of DA, i.e., \textit{why data augmentation works}. 
To facilitate this, we category DA methods according to the \textbf{diversity} of augmented data, since improving training data diversity is one of the main thrusts of DA effectiveness.
We frame DA methods into three categories, including paraphrasing, noising, and sampling.\footnote{This survey has been accepted by AI OPEN: \url{https://www.sciencedirect.com/journal/ai-open}. We also provide further resources at: \url{https://github.com/BohanLi0110/NLP-DA-Papers}}

Specifically, \textit{paraphrasing}-based methods generate the paraphrases of the original data as the augmented data. This category brings limited changes compared with the original data. \textit{Noising}-based methods add more continuous or discrete noises to the original data and involve more changes. \textit{Sampling}-based methods master the distribution of the original data to sample new data as augmented data. With the help of artificial heuristics and trained models, such methods can sample brand new data rather than changing existing data and therefore generate even more diverse data.

Our paper sets out to analyze DA methods in detail according to the above categories. In addition, we also introduce their applications in NLP tasks as well as the challenges.
The rest of the paper is structured as follows:
\begin{itemize}
\item Section~\ref{overview} presents a comprehensive review of the three categories and analyzes every single method in those categories. We also introduce the characteristics of the methods, e.g., the granularity and the level.

\item Section~\ref{tricks} refers to a summary of common strategies and tricks to improve the quality of augmented data, including method stacking, optimization, and filtering strategies.

\item Section~\ref{applications} analyzes the application of the above methods in NLP tasks. We also show the development of DA methods through a timeline.

\item Section~\ref{topics} introduces some related topics of data augmentation, including pre-trained language models, contrastive learning, similar data manipulation methods, generative adversarial networks, and adversarial attacks. We aim to connect data augmentation with other topics and show their difference at the same time.

\item Section~\ref{challenges} lists some challenges we observe in NLP data augmentation, including theoretical narrative and generalized methods. These points also reveal the future development direction of data augmentation.

\item Section~\ref{conclusion} concludes the paper.
\end{itemize}

\tikzstyle{leaf}=[%draw=hiddendraw,
    rounded corners,minimum height=1.2em,
    fill=myblue,text opacity=1,    align=center,
    fill opacity=.5,  text=black,align=left,font=\scriptsize,
inner xsep=3pt,
inner ysep=1pt,
]

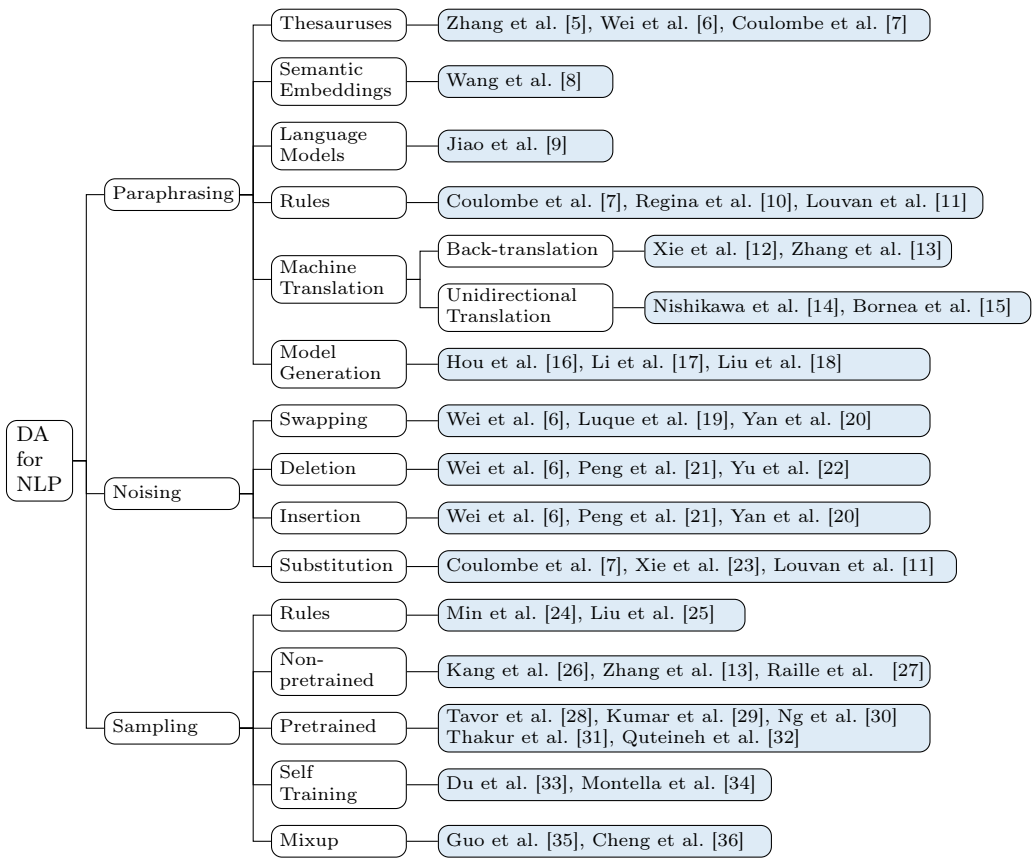
\begin{figure*}[thp]
  \centering
\begin{forest}
  forked edges,
  for tree={
  grow=east,
  reversed=true,%increase counter-clockwise
  anchor=base west,
  parent anchor=east,
  child anchor=west,
  base=left,
  font=\footnotesize,
  rectangle,
  draw=black,%hiddendraw,
  rounded corners,align=left,
  minimum width=2.5em,
  minimum height=1.2em,
  %  edge+={darkgray, line width=1pt},
%  l sep+=2.5pt,
%  s sep+=-5pt,
s sep=6pt,
inner xsep=3pt,
inner ysep=1pt,
  },
  %before packing={where n children=3{calign child=2, calign=child edge}{}},
  %before typesetting nodes={where content={}{coordinate}{}},
  %where level<=1{line width=2pt}{line width=1pt},
  where level=1{text width=4.5em,font=\scriptsize}{},
  where level=2{text width=4.5em,font=\scriptsize}{},
  where level=3{font=\scriptsize}{},
  where level=4{font=\scriptsize}{},
  where level=5{font=\scriptsize}{},
  [DA\\for\\NLP
    [Paraphrasing
        [Thesauruses
           [Zhang et al.~\cite{Zhang2015CharacterlevelCN[2]}{,} Wei et al.~\cite{Wei2019EDA[5]}{,} Coulombe et al.~\cite{Coulombe2018TextDA[12]}
           ,leaf,text width=18em]
        ]
        [Semantic\\Embeddings
            [Wang et al.~\cite{wang2015s[6]}
            ,leaf,text width=6em]
        ]
        [Language\\Models
            [Jiao et al.~\cite{Jiao2020TinyBERTDB[8]}
            ,leaf,text width=6em]    
        ]
        [Rules
            [Coulombe et al.~\cite{Coulombe2018TextDA[12]}{,} Regina et al.~\cite{Regina[40]}{,} Louvan et al.~\cite{louvan-magnini-2020-simple[43]}
             ,leaf,text width=20em]
        ]
        [Machine\\Translation
            [Back-translation, text width=6em
                [Xie et al.~\cite{xie2019unsupervised[9]}{,} Zhang et al.~\cite{Zhang2020ParallelDA[11]}
             ,leaf,text width=11em]
            ]
            [Unidirectional\\Translation, text width=6em
                [Nishikawa et al.~\cite{Nishikawa2020DataAW[31]}{,} Bornea et al.~\cite{Bornea2021MultilingualTL[48]}
             ,leaf,text width=14em]
            ]
        ]
        [Model\\Generation
            [Hou et al.~\cite{Hou2018SequencetoSequenceDA[18]}{,} Li et al.~\cite{li2020conditional[22]}{,} Liu et al.~\cite{Liu2020TellMH[30]}
            ,leaf,text width=18em]
        ]
    ]
    [Noising
        [Swapping
           [Wei et al.~\cite{Wei2019EDA[5]}{,} Luque et al.~\cite{Luque2019AtalayaAT[10]}{,} Yan et al.~\cite{Yan2019DataAF[15]}
           ,leaf,text width=18em]
        ]
        [Deletion
           [Wei et al.~\cite{Wei2019EDA[5]}{,} Peng et al.~\cite{Peng2020DataAF[37]}{,} Yu et al.~\cite{Yu2019HierarchicalDA[17]}
           ,leaf,text width=18em]
        ]
        [Insertion
           [Wei et al.~\cite{Wei2019EDA[5]}{,} Peng et al.~\cite{Peng2020DataAF[37]}{,} Yan et al.~\cite{Yan2019DataAF[15]}
           ,leaf,text width=18em]
        ]
        [Substitution
           [Coulombe et al.~\cite{Coulombe2018TextDA[12]}{,} Xie et al.~\cite{Xie2017DataNA[16]}{,} Louvan et al.~\cite{louvan-magnini-2020-simple[43]}
           ,leaf,text width=19em]
        ]
    ]
    [Sampling
        [Rules
            [Min et al.~\cite{Min2020SyntacticDA[13]}{,} Liu et al.~\cite{Liu2020ReverseOB[71]}
            ,leaf,text width=11em]
        ]
        [Non-\\pretrained
            [Kang et al.~\cite{Kang2018AdvEntuReAT[67]}{,} Zhang et al.~\cite{Zhang2020ParallelDA[11]}{,} Raille et al. ~\cite{Raille2020FastCD[54]}
                ,leaf,text width=18em]
        ]
        [Pretrained
            [Tavor et al.~\cite{anaby2019not[20]}{,} Kumar et al.~\cite{kumar2020data[21]}{,} Ng et al.~\cite{Ng2020SSMBASM[36]}\\ Thakur et al.~\cite{Thakur2021AugmentedSD[45]}{,} Quteineh et al.~\cite{quteineh-etal-2020-textual[38]}
            ,leaf,text width=18em]
        ]
        [Self\\Training
            [Du et al.~\cite{Du2021SelftrainingIP[69]}{,} Montella et al.~\cite{Montella2020DenoisingPA[47]}
            ,leaf,text width=12em]
        ]
        [Mixup,
        [Guo et al.~\cite{guo2019augmenting[24]}{,} Cheng et al.~\cite{cheng2020advaug[25]}
        ,leaf,text width=12em]
        ]
    ]
  ]
\end{forest}
\caption{Taxonomy of NLP DA methods.}
\label{Fig.taxonomy_of_DA}
\end{figure*}
\clearpage

\section{Data Augmentation Methods in NLP}
\label{overview}

\begin{figure}
    \centering
    \includegraphics[width=1\textwidth]{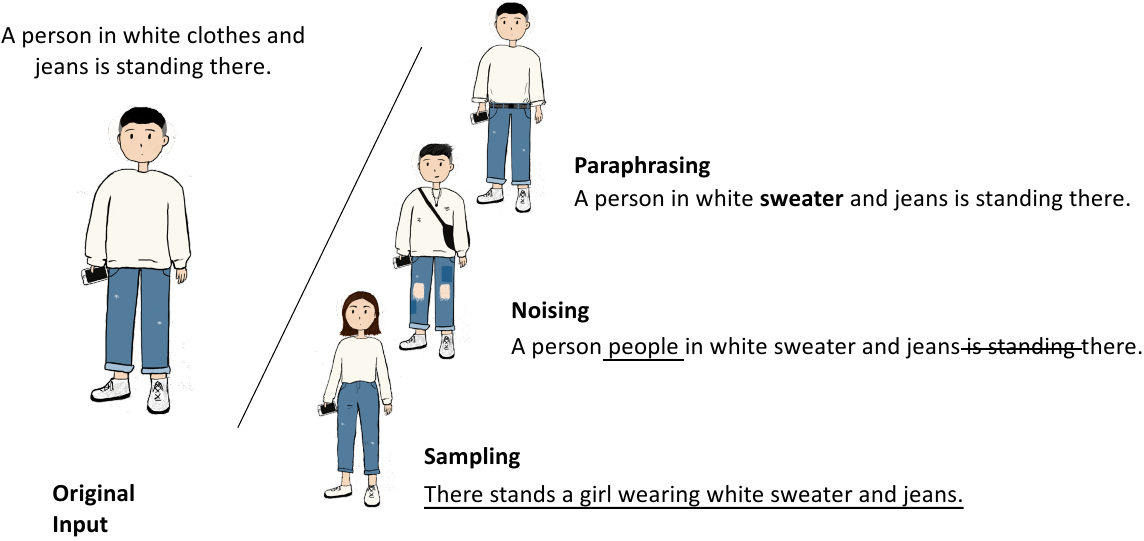}
    \caption{Data augmentation techniques include three categories. The examples of the original data and augmented data are on the left and right, respectively. As we can see, the \textbf{diversity} of \textit{paraphrasing}, \textit{noising}, and \textit{sampling} increases in turn compared to the original input.}
%In addition, all of them follow the meaning of the original input: ``A person in white clothes and jeans is standing there.
    \label{Fig.Three_types}
\end{figure}

Data Augmentation aims at generating additional, synthetic training data in insufficient data scenes. Data augmentation ranges from simple
techniques like rule-based methods to learnable generation-based methods, and all the above methods essentially guarantee the validity of the augmented data~\cite{Raille2020FastCD[54]}. That is to say, DA methods need to make sure that the augmented data is valid for the task, i.e.,  be considered to be part of the same distribution of the original data~\cite{Raille2020FastCD[54]}. For example, similar semantics in machine translation and the same label in text classification as the original data.

On the basis of validity, augmented data is also expected to be diverse to improve model generalization on downstream tasks. This involves the \textbf{diversity} of augmented data. In this survey, we novelly divide DA methods into three categories according to the diversity of their augmented data: paraphrasing, noising, and sampling.

\begin{itemize}
\item The paraphrasing-based methods generate augmented data that has limited semantic difference from the original data, based on proper and restrained changes to sentences. The augmented data convey very similar information as the original form.
\item The noising-based methods add discrete or continuous noise under the premise of guaranteeing validity. The point of such methods is to improve the robustness of the model.
\item The sampling-based methods master the data distributions and sample novel data within them. Such methods output more diverse data and satisfy more needs of downstream tasks based on artificial heuristics and trained models.
\end{itemize}

As shown in the examples and diagrams in Figure~\ref{Fig.Three_types}, the paraphrasing, noising, and sampling-based methods provide more diversity in turn. In this section, we will introduce and analyze them in detail.\footnote{The specific classification is shown in Figure~\ref{Fig.taxonomy_of_DA}}

\begin{figure}
    \centering
    \includegraphics[width=1\textwidth]{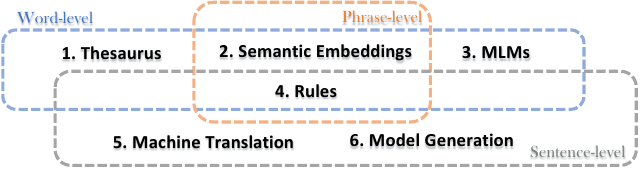}
    \caption{Data augmentation techniques by paraphrasing include three levels: word-level, phrase-level, and sentence-level.}
    \label{Fig.paraphrases}
\end{figure}

\subsection{\includegraphics[scale=0.02]{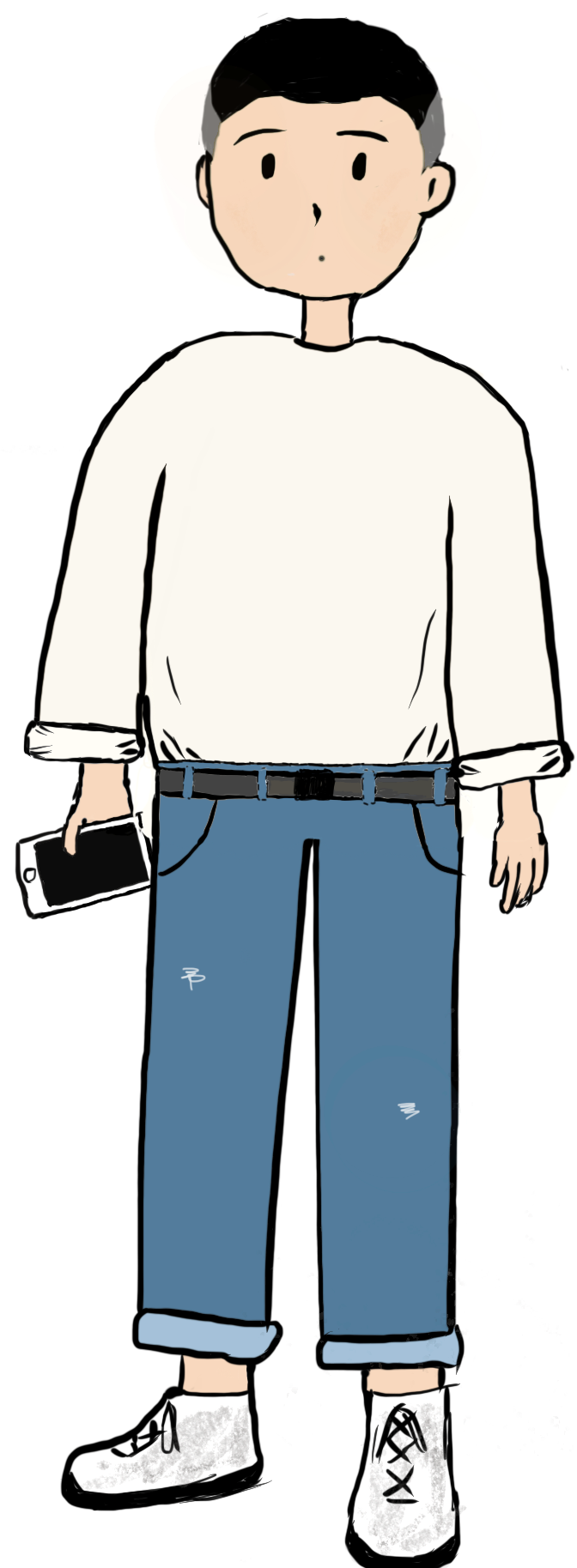} Paraphrasing-based Methods }

As common phenomena in natural language, paraphrases are alternative ways to convey the same information as the original form~\cite{barzilay2001extracting,madnani-dorr-2010-generating}. Naturally, the generation of paraphrases is a suitable scheme for data augmentation. Paraphrasing consists of several levels, including lexical paraphrasing, phrase paraphrase, and sentence paraphrase (Figure~\ref{Fig.paraphrases}). Therefore, the paraphrasing-based DA techniques introduced below can also be included into these three levels.

\subsubsection{Thesauruses}
\begin{figure}
    \centering
    \includegraphics[width=0.7\textwidth]{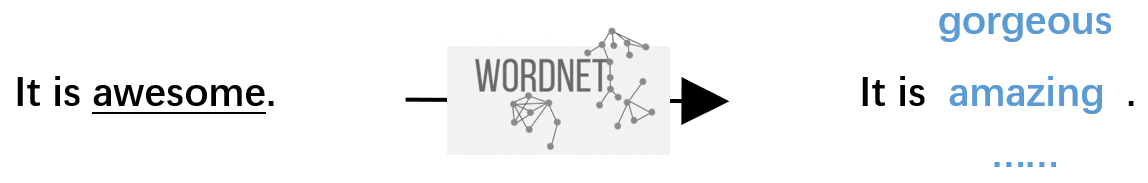}
    \caption{Paraphrasing by using thesauruses.}
    \label{Fig.paraphrasing_wordnet}
\end{figure}

Some works replace words in the original text with their synonyms and hypernyms,\footnote{Replacing a word with an antonym or a hyponym (more specific word) is usually not a semantically invariant transformation.~\cite{coulombe2018text[12]}} so as to obtain a new way of expression while keeping the semantics of the original text as unchanged as possible. As shown in Figure~\ref{Fig.paraphrasing_wordnet}, thesauruses like WordNet~\cite{Miller1995WordNetAL[4]} contain such lexical triplets of words and are often used as external resources.

Zhang et al.~\cite{Zhang2015CharacterlevelCN[2]} are the first to apply thesaurus in data augmentation. They use a thesaurus derived from WordNet,\footnote{The thesaurus is obtained from the Mytheas component used in LibreOffice project.} which sorts the synonyms of words according to their similarity. For each sentence, they retrieve all replaceable words and randomly choose $r$ of them to be replaced. The probability of number $r$ is determined by a geometric distribution with parameter $p$ in which $P [r]\sim p^{r}$. Given a word, the index $s$ of its chosen synonym is also determined by a another geometric distribution in which $P [s]\sim p^{s}$. The method ensures that synonyms that are more similar to the original word are selected with greater probability. Some methods~\cite{Mueller2016SiameseRA[3],daval-frerot-weis-2020-wmd[33],Dai2020AnAO[78]} apply a similar method.

A widely used text augmentation method called EDA (\textbf{E}asy \textbf{D}ata \textbf{A}ug-\\mentation Techniques)~\cite{Wei2019EDA[5]} also replaces the original words with their synonyms using WordNet: they randomly choose $n$ words, which are not stop words, from the original sentence.\footnote{$n$ is proportional to the length of the sentence.} Each of these words is replaced with a random synonym. Zhang et al.~\cite{Zhang2020OnDA[44]} apply a similar method in extreme multi-label classification.

In addition to synonyms, Coulombe et al.~\cite{Coulombe2018TextDA[12]} propose to use hypernyms to replace the original words. They also recommend the parts of speech of the augmented word in order of increasing difficulty: adverbs, adjectives, nouns, and verbs. Zuo et al.~\cite{Zuo2020KnowDisKE[96]} use WordNet and VerbNet~\cite{Schuler2005VerbnetAB} to retrieve synonyms, hypernyms, and words of the same category.

\tipbox{
\textbf{Thesauruses}

\small{
\textbf{Advantage(s):}

1. Easy to use.

\textbf{Limitation(s):}

1. The scope and part of speech of augmented words are limited.

2. This method cannot resolve the ambiguity problem.

3. Sentence semantics may be affected if there are too many substitutions.
}}

\subsubsection{Semantic Embeddings}
\begin{figure}
    \centering
    \includegraphics[width=0.7\textwidth]{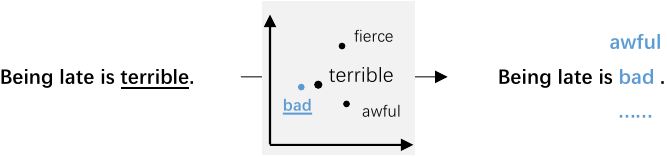}
    \caption{Paraphrasing by using semantic embeddings.}
    \label{Fig.paraphrasing_embeddings}
\end{figure}

This method overcomes the limitations of replacement range and parts of speech in the thesaurus-based method. It uses pre-trained word embeddings, such as Glove, Word2Vec, FastText, etc., and replaces the original word in the sentence with its closest neighbor in embedding space, as shown in Figure~\ref{Fig.paraphrasing_embeddings}.

In the Twitter message classification task, Wang et al.~\cite{wang2015s[6]} pioneer to use both word embeddings and frame embeddings instead of discrete words.\footnote{The frame embeddings refer to the continuous embeddings of semantic frames~\cite{baker-etal-1998-berkeley-framenet}.} As for word embeddings, each original word in the tweet is replaced with one of its k-nearest-neighbor words using cosine similarity. For example, ``Being late is terrible'' becomes ``Being behind are bad''. As for frame semantic embeddings, the authors semantically parse 3.8 million tweets and build a continuous bag-of-frame model to represent each semantic frame using Word2Vec~\cite{mikolov2013word2vec}. The same data augmentation approach as words is then applied to semantic frames.

Compared to Wang et al.~\cite{wang2015s[6]}, Liu et al.~\cite{Liu2020DocumentlevelMS[7]} only use word embeddings to retrieve synonyms. In the meanwhile, they edit the retrieving result with a thesaurus for balance. RamirezEchavarria et al.~\cite{RamirezEchavarria2020OnTE[70]} create the dictionary of embeddings for selection.

\tipbox{
\textbf{Semantic Embeddings}

\small{
\textbf{Advantage(s):}

1. Easy to use.

2. Higher replacement hit rate and more comprehensive replacement range.

\textbf{Limitation(s):} 

1. This method cannot resolve the ambiguity problem.\footnote{Static word embeddings such as Wrod2Vec have only one representation for each word.}

2. Sentence semantics may be affected if there are too many substitutions.}
}

\subsubsection{Language Models}
\begin{figure}
    \centering
    \includegraphics[width=0.7\textwidth]{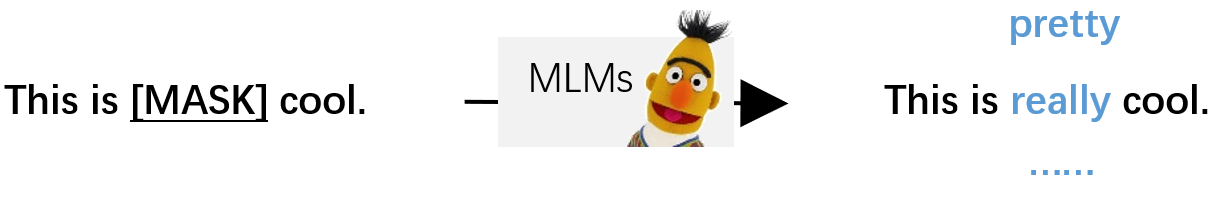}
    \caption{Paraphrasing by using language models.}
    \label{Fig.paraphrasing_MLMs}
\end{figure}

Pretrained language models have become mainstream models in recent years due to their excellent performance. Masked language models (MLMs) such as BERT and RoBERTa can predict masked words in text based on context, which can be used for text data augmentation (as shown in Figure~\ref{Fig.paraphrasing_MLMs}). Moreover, this approach alleviates the ambiguity problem since MLMs consider the whole context.

Wu et al.~\cite{Wu2019ConditionalBC[243]} fine-tune on pre-trained BERT to perform conditional MLM task. They alter the segmentation embeddings to label embeddings, which are learned corresponding to the annotated labels on labeled datasets. They use this fine-tuned conditional BERT to augment sentences. Specifically, a few words in a labeled sentence are randomly mask then filled by the conditional BERT. %%有一个前人工作可以放在这里  Contextual augmentation: Data augmentation by words with paradigmatic relations。是kobayashi-2018-contextual[174] 换一下这一段的结构，并且调整两者的顺序

Jiao et al.~\cite{Jiao2020TinyBERTDB[8]} use both word embeddings and masked language models to obtain augmented data. They apply the tokenizer of BERT to tokenize words into multiple
word pieces. Each word piece is replaced with probability 0.4. If a word piece is not a complete word (``est'' for example), it is replaced by its K-nearest-neighbor words in the Glove embedding space. If the word piece is a complete word, the authors replace it with [MASK] and employ BERT to predict $K$ Words to fill in the blank. 
Regina et al.~\cite{Regina[40]}, Tapia-T{\'e}llez et al.~\cite{TapiaTllez2020DataAW[49]}, Lowell et al.~\cite{Lowell2020UnsupervisedDA[61]}, and Palomino et al.~\cite{Palomino2020PalominoOchoaAT[51]} apply methods that are similar to Jiao et al.~\cite{Jiao2020TinyBERTDB[8]}. They mask multiple words in a sentence and generate new sentences by filling these masks to generate more varied sentences. 
In addition, RNNs are also used for replacing the original word based on the context (\cite{kobayashi-2018-contextual[174],fadaee-etal-2017-data[183]}).

\tipbox{
\textbf{Language Models}

\small{
\textbf{Advantage(s):}

1. This approach alleviates the ambiguity problem.

2. This method considers context semantics.

\textbf{Limitation(s):} 

1. Still limited to the word level.

2. Sentence semantics may be affected if there are too many substitutions.
}}

\subsubsection{Rules}
\begin{figure}
    \centering
    \includegraphics[width=0.7\textwidth]{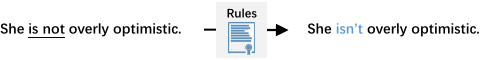}
    \caption{Paraphrasing by using rules.}
    \label{Fig.paraphrasing_rules}
\end{figure}

This method requires some heuristics about natural language to ensure the maintaining of sentence semantics, as shown in Figure~\ref{Fig.paraphrasing_rules}.

On the one hand, some works rely on existing dictionaries or fixed heuristics to generate word-level and phrase-level paraphrases. Coulombe et al.~\cite{Coulombe2018TextDA[12]} introduce the use of regular expressions to transform the form without changing sentence semantics, such as the abbreviations and prototypes of verbs, modal verbs, and negation. For example, replace ``is not'' with ``isn't''. Similarly, Regina et al.~\cite{Regina[40]} use word-pair dictionaries to perform replacements between the expanded form and the abbreviated form.

On the other hand, some works generate sentence-level paraphrases for original sentences with some rules, e.g. dependency trees. 
Coulombe et al.~\cite{Coulombe2018TextDA[12]} use a syntactic parser to build a dependency tree for the original sentence. Then the dependency tree is used for syntax transformation. For example, replace ``Sally embraced Peter excitedly.'' with ``Peter was embraced excitedly by Sally.''. Dehouck et al.~\cite{Dehouck2020DataAV[101]} apply a similar method. Louvan et al.~\cite{louvan-magnini-2020-simple[43]} crop particular fragments on the dependency tree to create a smaller sentence. They also rotate the target fragment around the root of the dependency parse structure, without harming the original semantics.

\tipbox{
\textbf{Rules}

\small{
\textbf{Advantage(s):}

1. Easy to use.

2. This method preserves the original sentence semantics.

\textbf{Limitation(s):} 

1. This method requires artificial heuristics.

2. Low coverage and limited variation.
}}

\subsubsection{Machine Translation}
\begin{figure}
    \centering
    \includegraphics[width=0.7\textwidth]{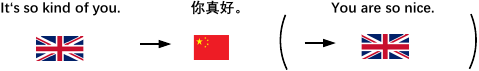}
    \caption{Paraphrasing by machine translation.}
    \label{Fig.paraphrasing_MT}
\end{figure}

Translation is a natural means of paraphrasing. With the development of machine translation models and the availability of online APIs, machine translation is popular as an augmentation method in many tasks, as shown in Figure~\ref{Fig.paraphrasing_MT}.
\paragraph{\textbf{Back-translation}}
\label{back-translation}
This method means that the original text is translated into other languages, and then translated back to obtain the augmented text in the original language. Different from word-level methods, back-translation does not directly replace individual words but rewrites the whole sentence in a generated way.

Xie et al.~\cite{xie2019unsupervised[9]}, Yu et al.~\cite{Yu2018QANetCL[249]}, and Fabbri et al.~\cite{Fabbri2021ImprovingZA[77]} use English-French translation models (in both directions) to perform back-translation on each sentence and obtain their paraphrases. Lowell et al.~\cite{Lowell2020UnsupervisedDA[61]} also introduce this method as one of the unsupervised data augmentation methods. Zhang et al.~\cite{Zhang2020ParallelDA[11]} leverage back-translation to obtain the formal expression of the original data in the style transfer task. 

In addition to some trained machine translation models, some cloud translation API services like Google and DeepL are common tools for back-translation and are applied by some works like ~\cite{Coulombe2018TextDA[12],Luque2019AtalayaAT[10],Ibrahim2020AlexUBackTranslationTLAS[32],daval-frerot-weis-2020-wmd[33],longpre-etal-2020-effective[34],Rastogi2020CanWA[39],Regina[40],Aleksandr-inproceedings[50],aroyehun-gelbukh-2018-aggression[165]}.\footnote{The links of the above Cloud Translation API services are: \url{https://cloud.google.com/translate/docs/apis} (Google) and \url{https://www.deepl.com/translator} (DeepL).}

Some works add additional features based on vanilla back-translation. Nugent et al.~\cite{Nugent2020DetectingET[46]} propose a range of softmax temperature settings to ensure diversity while preserving semantic meaning. Qu et al.~\cite{Qu2021CoDACA[64]} combine back-translation with adversarial training, to synthesize diverse and informative augmented examples by organically integrating multiple transformations. Zhang et al.~\cite{Zhang2020ParallelDA[11]} employ a discriminator to filter the sentences in the back-translation results. This method greatly improves the quality of the augmented data as a threshold.

\paragraph{\textbf{Unidirectional Translation}}
Different from back-translation, the unidirectional translation method directly translates the original text into other languages once, without translating it back to the original language. This method usually occurs in a multilingual scene.

In the task of unsupervised cross-lingual word embeddings (CLWEs), Nishikawa et al.~\cite{Nishikawa2020DataAW[31]} build pseudo-parallel corpus with an unsupervised machine translation model. The authors first train unsupervised machine translation (UMT) models using the source/target training corpora and then translate the corpora using the UMT models. The machine-translated corpus is used together with the original corpus to learn monolingual word embeddings for each language independently. Finally, the learned monolingual word embeddings are mapped to a shared CLWE space. This method both facilitates the structural similarity of two monolingual embedding spaces and improves the quality of CLWEs in the unsupervised mapping method.

Bornea et al.~\cite{Bornea2021MultilingualTL[48]}, Barrire et al.~\cite{Barrire2020ImprovingSA[89]}, and Aleksandr et al.~\cite{Aleksandr-inproceedings[50]} translate the original English corpus into several other languages and obtain multiplied data. Correspondingly, they use multilingual models.

\tipbox{
\textbf{Machine Translation}

\small{
\textbf{Advantage(s):}

1. Easy to use.

2. Wide range of applications.

3. This approach guarantees correct syntax and unchanged semantics.

\textbf{Limitation(s):} 

1. Poor controllability and limited diversity because of the fixed machine translation models.
}}

\subsubsection{Model Generation}
\begin{figure}
    \centering
    \includegraphics[width=0.7\textwidth]{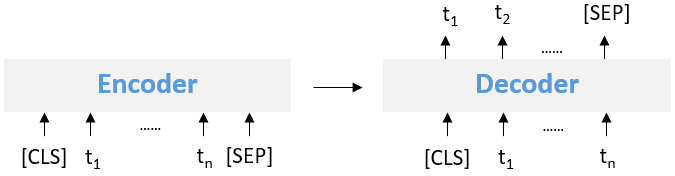}
    \caption{Paraphrasing by model generation.}
    \label{Fig.paraphrasing_generation}
\end{figure}

Some methods employ Seq2Seq models to generate paraphrases directly. Such models output more diverse sentences given proper training objects, as shown in Figure~\ref{Fig.paraphrasing_generation}.

Hou et al.~\cite{Hou2018SequencetoSequenceDA[18]} propose a Seq2Seq data augmentation model for the language understanding module of task-based dialogue systems. They feed the delexicalized input utterance and the specified diverse rank $k$ (e.g. 1, 2, 3) into the Seq2Seq model as the input to generate a new utterance.
Similarly, Hou et al.~\cite{Hou2020C2CGenDACG[19]} encodes the concatenated multiple input utterances by an L-layer transformer. The proposed model uses duplication-aware attention and diverse-oriented regularization to generate more diverse sentences.

In the task of aspect term extraction, Li et al.~\cite{li2020conditional[22]} adopt Transformer as the basic structure. The masked original sentences as well as their label sequences are used to train a model $M$ that reconstructs the masked fragment as the augmented data.\footnote{Half of the words in original sentences whose sequence labels are not `O' are masked.} Kober et al.~\cite{Kober2021DataAF[97]} use GAN to generate samples that are very similar to the original data.
Liu et al.~\cite{Liu2020TellMH[30]} employ a pre-trained model to provide prior information to the proposed Transformer-based model. Then the proposed model could generate both context-relevant answerable questions and unanswerable questions.

\tipbox{
\textbf{Model Generation}

\small{
\textbf{Advantage(s):}

1. Wide range of applications.

2. Strong application.

\textbf{Limitation(s):} 

1. Require for training data.

2. High training difficulty.
}}

\subsection{\includegraphics[scale=0.02]{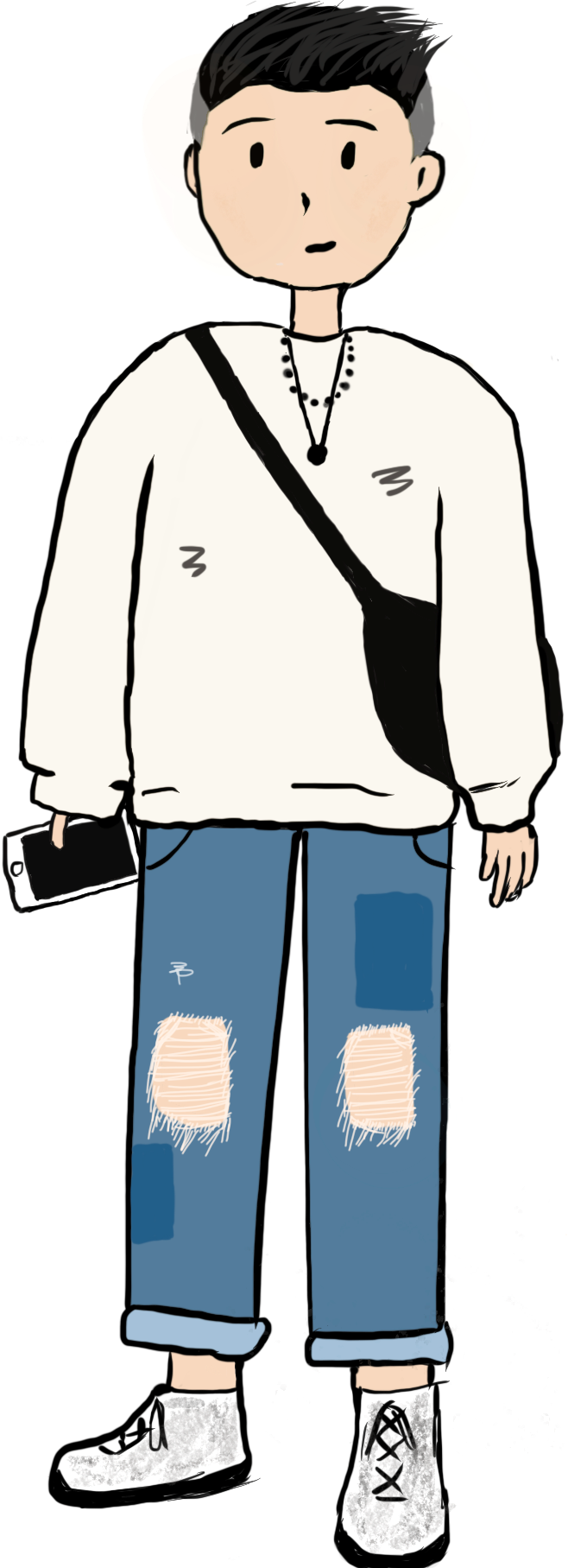} Noising-based Methods}

The focus of paraphrasing is to make the semantics of the augmented data as similar to the original data as possible. In contrast, the noising-based methods add faint noise that does not seriously affect the semantics, so as to make it appropriately deviate from the original data. Humans greatly reduce the impact of weak noise on semantic understanding through their grasp of linguistic phenomena and prior knowledge, but this noise can pose challenges for models. Thus, this method not only expands the amount of training data but also improves model robustness.

\begin{figure}
    \centering
    \includegraphics[width=0.8\textwidth]{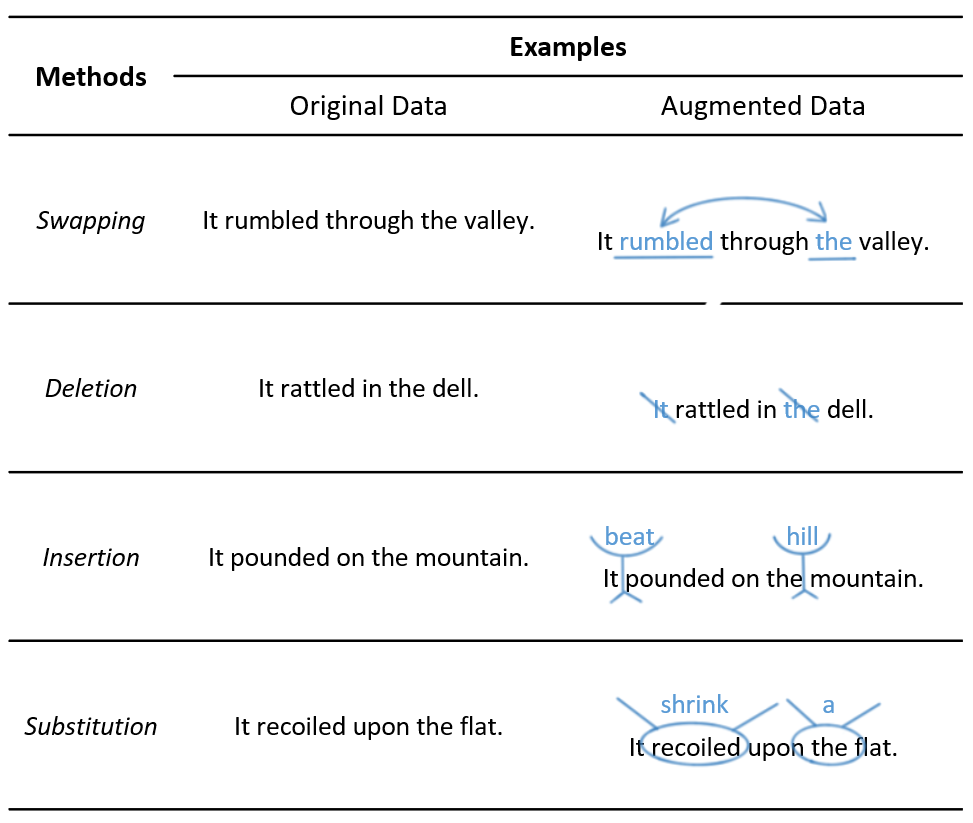}
    \caption{The example of five noising-based methods.}
    \label{Fig.Noising_methods}
\end{figure}

\subsubsection{Swapping}
The semantics of natural language are sensitive to text order, while slight order change is still readable for humans~\cite{wang1999reading}. Therefore, the random swapping between words even sentences within a reasonable range can be used as a data augmentation method.

Wei et al.~\cite{Wei2019EDA[5]} randomly choose two words in the sentence and swap their positions. This process is repeated $n$ times, in which $n$ is proportional to the sentence length $l$. Longpre et al.~\cite{longpre-etal-2020-effective[34]}, Rastogi et al.~\cite{Rastogi2020CanWA[39]}, and Zhang et al.~\cite{Zhang2020OnDA[44]} also apply the same method. Dai et al.~\cite{Dai2020AnAO[78]} split the token sequence into segments according to labels, then randomly choose some segments to shuffle the order of the tokens inside, with the label order unchanged.

In addition to word-level swapping, some works also propose sentence-level even instance-level swapping. In the task of tweet sentiment analysis, Luque et al.~\cite{Luque2019AtalayaAT[10]} divide tweets into two halves. They randomly sample and combine first halves with second halves that have the same label. Although the data generated in this way may be ungrammatical and semantically unsound, it still carries relatively complete semantics and emotional polarity compared to individual words. Yan et al.~\cite{Yan2019DataAF[15]} perform sentence-level random swapping on legal documents classification. Since sentences independently contain relatively complete semantics comparing to words, the sentence order in the legal document has little effect on the meaning of the original text. Consequently, the authors shuffle the sentences to obtain the augmented text.

\subsubsection{Deletion}
This method means randomly deleting words in a sentence or deleting sentences in a document.

As for word-level deletion, Wei et al.~\cite{Wei2019EDA[5]} randomly remove each word in the sentence with probability $p$. Longpre et al.~\cite{longpre-etal-2020-effective[34]}, Rastogi et al.~\cite{Rastogi2020CanWA[39]}, and Zhang et al.~\cite{Zhang2020OnDA[44]} also apply the same method. In the task of spoken language understanding, Peng et al.~\cite{Peng2020DataAF[37]} augment input dialogue acts by deleting slot values to obtain more combinations.

As for sentence-level deletion, Yan et al.~\cite{Yan2019DataAF[15]} randomly delete each sentence in a legal document according to a certain probability. They do this because there exist many irrelevant statements and deleting them will not affect the understanding of the legal case. Yu et al.~\cite{Yu2019HierarchicalDA[17]} employ the attention mechanism to determine the objective of both word-level and sentence-level random deletion.

\subsubsection{Insertion}
This method means randomly inserting words into a sentence or inserting sentences into a document.

As for word-level insertion, Wei et al.~\cite{Wei2019EDA[5]} select a random synonym of a random word in a sentence that is not a stop word, then insert that synonym into a random position in the sentence. This process is repeated $n$ times. In the task of spoken language understanding, Peng et al.~\cite{Peng2020DataAF[37]} augment input dialogue acts by inserting slot values to obtain more combinations.

In legal documents classification, since documents with the same label may have similar sentences, Yan et al.~\cite{Yan2019DataAF[15]} employ sentence-level random insertion. They randomly select sentences from other legal documents with the same label to get augmented data.

\notebox{\small{
Random insertion introduces new noisy information that may change the original label. Tips to avoid this problem:

1. Word level: use label-independent external resources.

2. Sentence level: use other samples with the same labels as the original data.
}}

\subsubsection{Substitution}
This method means randomly replacing words or sentences with other strings. Different from the above paraphrasing methods, this method usually avoids using strings that are semantically similar to the original data.

Some works implement substitution through existing outer resources. Coulombe et al.~\cite{Coulombe2018TextDA[12]} and Regina et al.~\cite{Regina[40]} introduce a list of the most common misspellings in English to generate augmented texts containing common misspellings.\footnote{A list of common spelling errors in English can be obtained from the online resources of Oxford Dictionaries: \url{https://en.oxforddictionaries.com/spelling/common-misspellings}} For example, ``across'' is easily misspelled as ``accross''.
Xie et al.~\cite{Xie2017DataNA[16]} borrow from the idea of ``word-dropout'' and improve generalization by reducing the information in the sentence. This work uses ``\_'' as a placeholder to replace random words, indicating that the information at that position is empty.% Peng et al. ~\cite{Peng2020DictionarybasedDA[53]} use pseudo-IND parallel corpus embeddings to create dictionaries and generate augmented data.

Some works use task-related resources or generate random strings for substitution. Xie et al.~\cite{xie2019unsupervised[9]} and Xie et al.~\cite{Xie2017DataNA[16]} replace the original words with other words in the vocabulary, and they use the TF-IDF value and the unigram frequency to choose words from the vocabulary, respectively. Lowell et al.~\cite{Lowell2020UnsupervisedDA[61]} and Daval et al.~\cite{daval-frerot-weis-2020-wmd[33]} also explore this method as one of unsupervised data augmentation methods. Wang et al.~\cite{wang-etal-2018-switchout[170]} propose a method that randomly replaces words in the input and target sentences with other words in the vocabulary. In NER, Dai et al.~\cite{Dai2020AnAO[78]} replace the original token with a random token in the training set with the same label.
Qin et al.~\cite{Qin2020CoSDAMLMC[103]} propose a multi-lingual code-switching method that replaces original words in the source language with words of other languages.
In the task of task-oriented dialogue, random substitution is a useful way to generate augmented data. Peng et al.~\cite{Peng2020DataAF[37]} augment input dialogue acts by replacing slot values to obtain more combinations in spoken language understanding.
In slot filling, Louvan et al.~\cite{louvan-magnini-2020-simple[43]} do slot substitution according to the slot label. Song et al.~\cite{Song2021DataAF[52]} augment the training data for dialogue state tracking by copying user utterances and replace the corresponding real slot values with generated random strings.

\notebox{\small{
Random substitution introduces new noisy information that may change the original label. Tips to avoid this problem:

1. Word level: use label-independent external resources.

2. Sentence level: use other samples with the same labels as the original data.
}}

\tipbox{
\textbf{Noising}

\small{
\textbf{Advantage(s):}

1. Noising-based methods improve model robustness.
}

\small{
\textbf{Disadvantage(s):}

1. Poor interpretability.

2. Limited diversity for every single method.
}}

\subsection{\includegraphics[scale=0.02]{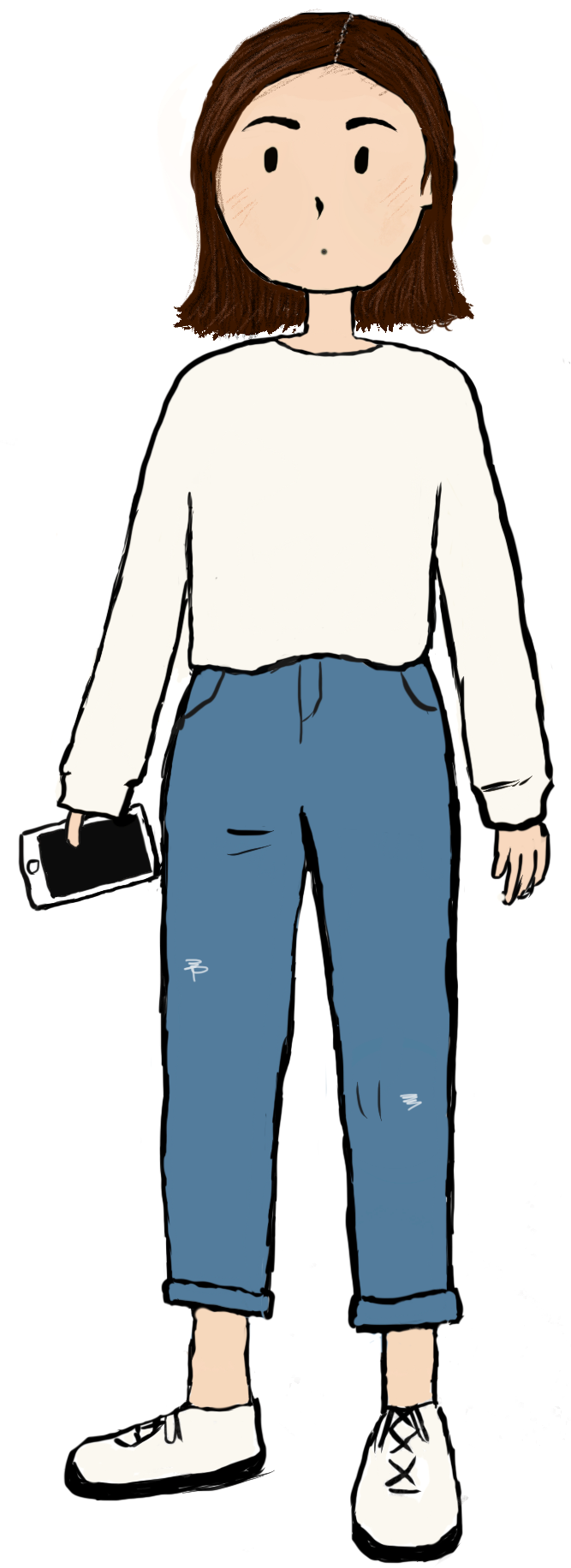} Sampling-based Methods}

Sampling-based methods grasp the data distribution and sample new data within it. Similar to paraphrasing-based models, they also involve rules and trained models to generate augmented data. The difference is that the sampling-based methods are task-specific and require task information like labels and data format.\footnote{Recall that paraphrasing-based methods are task-independent and only require the original sentence as input.} Such methods not only ensure validity but also increase diversity. They satisfy more needs of downstream tasks based on artificial heuristics and trained models, and can be designed according to specific task requirements. Thus, they are usually more flexible and difficult than the former two categories.

\subsubsection{Rules}
\begin{figure}
    \centering
    \includegraphics[width=1\textwidth]{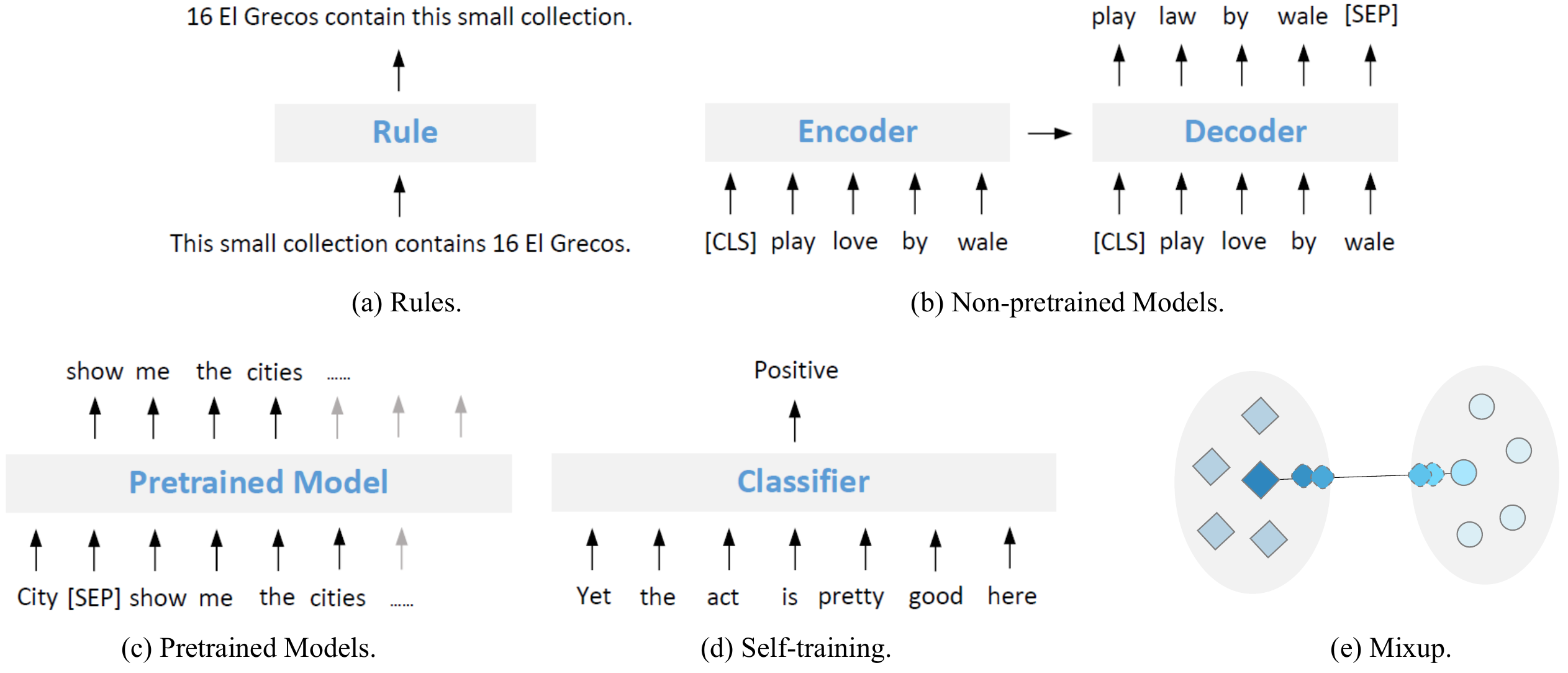}
    \caption{Sampling-based models.}
    \label{Fig.sampling-based}
\end{figure}

This method uses some rules to directly generate new augmented data. Heuristics about natural language and the corresponding labels are sometimes required to ensure the validity of the augmented data. The model structure is as shown in Figure~\ref{Fig.sampling-based}(a). Different from the above rule-based paraphrasing method, this method constructs valid but not guaranteed to be similar to the original data (even different labels).

Min et al.~\cite{Min2020SyntacticDA[13]} swap the subject and object of the original sentence, and convert predicate verbs into passive form. For example, inverse ``This small collection contains 16 El Grecos.'' into ``16 El Grecos contain this small collection.''. The labels of new samples are determined by rules.
Liu et al.~\cite{Liu2020ReverseOB[71]} apply data augmentation methods in the task of solving math word problems (MWPs). They filter out some irrelevant numbers. Then some rules are used to construct new data based on the idea of double-checking, e.g., constructing augmented data describing $distance = time \times speed$ by reusing the original data describing $time = distance / speed$. The output equations of this method are computationally right.
Given the training set of Audio-Video Scene-Aware Dialogue that provides 10 question-answer pairs for each video, Mou et al.~\cite{Mou2020MultimodalDS[42]} shuffle the first $n$ pairs as dialogue history and take the $n + 1$-th question as what needs to be answered.
In natural language inference, Kang et al.~\cite{Kang2018AdvEntuReAT[67]} apply external resources like PPDB and artificial heuristics to construct new sentences. Then they combine the new sentences with original sentences as augmented pairs according to rules, for example, \textit{if A entails B and B entails C, then A entails C}.
Kober et al.~\cite{Kober2021DataAF[97]} define some rules to construct positive and negative pairs using adjective-noun (AN) and noun-noun (NN) compounds. For example, given $<car,car>$, they construct $<fast car,car>$ as a positive sample and $<fast car,red car>$ as a negative sample.
Shakeel et al.~\cite{Shakeel2020AMM[128]} construct both paraphrase annotations and non-paraphrase annotations through three properties including reflexivity, symmetry, and transitive extension.
Yin et al.~\cite{Yin2020DialogST[143]} use two kinds of rules including symmetric consistency and transitive consistency, as well as logic-guided DA methods to generate DA samples.

\tipbox{
\textbf{Rules}

\small{
\textbf{Advantage(s):}

1. Easy to use.

\textbf{Limitation(s):} 

1. Require for artificial heuristics.

2. Low coverage and limited variation.
}}

\subsubsection{Non-pretrained Models}
Some methods use non-pretrained models to generate augmented data. Such methods usually entail the idea of \textbf{back translation (BT)}~\cite{sennrich-etal-2016-improving},\footnote{Note that the idea of back translation here is DIFFERENT from the above paraphrasing method called ``back-translation'' in Section~\ref{back-translation}.} which is to train a target-to-source Seq2Seq model and use the model to generate source sentences from target sentences, i.e., constructing pseudo-parallel sentences~\cite{Zhang2020ParallelDA[11]}. Such Seq2Seq model learns the internal mapping between the distributions of the target and the source, as shown in Figure~\ref{Fig.sampling-based}(b). This is different from the model generation based paraphrasing method because the augmented data of the paraphrasing method shares similar semantics with the original data.

Sennrich et al.~\cite{Sennrich2016ImprovingNM[244]} train an English-to-Chinese NMT model using existing parallel corpus, and use the target English monolingual corpus to generate Chinese corpus through the above English-to-Chinese model.
Kang et al.~\cite{Kang2018AdvEntuReAT[67]} train a Seq2Seq model for each label ($entailment$, $contradiction$, and $neutral$) and then generate new data using the Seq2Seq model given a sentence and a specific label. 
Chen et al.~\cite{Chen2020PatternawareDA[65]} adopt the Tranformer architecture and map the ``rewrite utterance $\rightarrow$ request utterance'' to the machine translation process. Moreover, they enforce the optimization process of the Seq2Seq generation with a policy gradient technique for controllable rewarding. Zhang et al.~\cite{Zhang2020ParallelDA[11]} use Transformer as the encoder and transfer the knowledge from Grammatical Error Correction to Formality Style Transfer.
Raille et al.~\cite{Raille2020FastCD[54]} create the Edit-transformer, a Transformer-based model works cross-domain. Yoo et al.~\cite{Yoo2019DataAF[161]} propose a novel VAE model to output the semantic slot sequence and the intent label given an utterance.

\tipbox{
\textbf{Non-pretrained Models}

\small{
\textbf{Advantage(s):}

1. Strong diversity.

2. Strong application.

\textbf{Limitation(s):} 

1. Require training data.

2. High training difficulty.
}}

\subsubsection{Pretrained Models}
In recent years, large-scale language models (LM) have achieved great success by acquiring rich linguistic knowledge through pretraining. Thus, they are naturally used as augmentation tools, as shown in Figure~\ref{Fig.sampling-based}(c).

Tavor et al.~\cite{anaby2019not[20]} propose a data augmentation method named LAMBDA. They generate labeled augmented sentences with GPT-2, which is fine-tuned on the training set in advance. Then the augmented sentences are filtered by a classifier to ensure the data quality. Kumar et al.~\cite{kumar2020data[21]} applies a similar method without the classifier for filtering.

Some works adopt masked language models to obtain augmented data. Ng et al.~\cite{Ng2020SSMBASM[36]} use the masked language model to construct a corruption model and a reconstruction model. Given the input data points, they initially generate data far away from the original data manifold with the corruption model. Then the reconstruction model is used to pull the data point back to the original data manifold as the final augmented data.

Some works adopt auto-regressive models to obtain augmented data.
Peng et al.~\cite{Peng2020DataAF[37]} use the pre-trained SC-GPT and SC-GPT-NLU to generate utterances and dialogue acts respectively. The results are filtered to ensure the data quality. Abonizio et al.~\cite{Abonizio2020PretrainedDA[93]} fine-tune DistilBERT~\cite{Sanh2019DistilBERTAD} on original sentences to generate synthetic sentences. Especially, GPT-2 is a popular model used for generating augmented data. Quteineh et al.~\cite{quteineh-etal-2020-textual[38]} use label-conditioned GPT-2 to generate augmented data. Tarj{\'a}n et al.~\cite{tarjan2020deep[41]} generate augmented data with GPT-2 and retokenize them into statistically derived subwords to avoid the vocabulary explosion in a morphologically rich language. Zhang et al.~\cite{Zhang2020OnDA[44]} use GPT-2 to generate substantially diversified augmented data in extreme multi-label classification. 

\tipbox{
\textbf{Pretrained Models}

\small{
\textbf{Advantage(s):}

1. Strong diversity.

2. Strong application.

\textbf{Limitation(s):} 

1. Require training data.
}}

\subsubsection{Self-training}
In some scenarios, unlabeled raw data is easy to obtain. Thus, converting such data into valid data would greatly increase the amount of data, as shown in Figure~\ref{Fig.sampling-based}(d).

Thakur et al.~\cite{Thakur2021AugmentedSD[45]} first fine-tune BERT on the original data, then use the fine-tuned BERT to label unlabeled sentence pairs. Such augmented data, as well as the gold data, are used to train SBERT together.
Miao et al.~\cite{Miao2020TwitterDA[35]} further introduce data distillation into the self-training process. They output the label of unlabeled data by the iteratively updated teacher model. Yang et al.~\cite{Yang2020NeuralRF[59]} apply a similar self-training method in question answering; a cross-attention-based teacher model is used to determine the label of each QA pair. Du et al.~\cite{Du2021SelftrainingIP[69]} introduce SentAugment, a data augmentation method that computes task-specific query embeddings from labeled data to retrieve sentences from a bank of billions of unlabeled sentences crawled from the web.

Some methods directly transfer exsiting models from other tasks to generate pseudo-parallel corpus.
Montella et al.~\cite{Montella2020DenoisingPA[47]} make use of Wikipedia to leverage a massive sentences. Then they use Stanford OpenIE package to extract the triplets given Wikipedia sentences. For example, given ``\textit{Barack Obama was born in Hawaii.}'', the returned triples by Stanford OpenIE are $<Barack Obama; was; born> and <Barack Obama; was born in; Hawaii>$ Such mappings are flipped as the augmented data of RDF-to-text tasks. Aleksandr et al.~\cite{Aleksandr-inproceedings[50]} apply a similar method. Since BERT does well on object-property (OP) relationship prediction and object-affordance (OA) relationship prediction, Zhao et al.~\cite{Zhao2020LearningPC[86]} directly use a fine-tuned BERT to predict the label of OP and OA samples.

\tipbox{
\textbf{Self-training}

\small{
\textbf{Advantage(s):}

1. Easier than generative models.

2. Suitable for data-sparse scenarios.
}

\small{
\textbf{Disadvantage(s):}

1. Require for unlabeled data.
}}

\subsubsection{Mixup}
This method uses virtual embeddings instead of generated natural language form text as augmented samples. The existing data is used as the basis to sample in the virtual vector space, and the sampled data may have different labels than the original data.

The idea of Mixup first appears in the image field by Zhang et al.~\cite{zhang2017mixup[23]}. Inspired by this work, Guo et al.~\cite{guo2019augmenting[24]} propose two variants of Mixup for sentence classification. The first one called wordMixup conducts sample interpolation in the word embedding space, and the second one called senMixup interpolates the hidden states of sentence encoders. The interpolated new sample through wordMixup as well as senMixup, and their common interpolated label are obtained as follows:
\begin{equation}
\label{mixup_wordMixup}
\widetilde{B}_{t}^{i j}=\lambda B_{t}^{i}+(1-\lambda) B_{t}^{j},
\end{equation}
\begin{equation}
\label{mixup_senMixup}
\widetilde{B}_{\{k\}}^{i j}=\lambda f( B^{i})_{\{k\}}+(1-\lambda) f(B^{j})_{\{k\}},
\end{equation}
\begin{equation}
\label{mixup_labelMixup}
\tilde{y}^{i j}=\lambda y^{i}+(1-\lambda) y^{j},
\end{equation}
in which $B_{t}^{i}, B_{t}^{j} \in R^{N \times d}$ denote the $t$-th word in two original sentences, and $f(B^i), f(B^j)$ denote the hidden layer sentence representation. Moreover, $y^i, y^j$ are the corresponding original labels.

Mixup is widely applied in many works recently. Given the original samples, Cheng et al.~\cite{cheng2020advaug[25]} firstly construct their adversarial samples following \cite{cheng-etal-2019-robust}, and then apply two Mixup strategies named $P_{adv}$ and $P_{aut}$: The former interpolates between adversarial samples, and the latter interpolates between the two corresponding original samples. Similarly, Sun et al.~\cite{Sun2020MixupTransformerDD[75]}, Bari et al.~\cite{BARI2020MultiMixAR[55]} , and Si et al.~\cite{Si2020BetterRB[83]} both apply such Mixup method for text classification. Sun et al.~\cite{Sun2020MixupTransformerDD[75]} propose Mixup-Transformer which combines Mixup with transformer-based pre-trained architecture. They test its performance on text classification datasets. Chen et al.~\cite{Chen2020LocalAB[72]} introduce Mixup into NER, proposing both Intra-LADA and InterLADA.

\tipbox{
\textbf{Mixup}

\small{
\textbf{Advantage(s):}

1. Generating augmented data between different labels.

}

\small{
\textbf{Disadvantage(s):}

1. Poor interpretability.
}}

\subsection{Analysis}
As shown in Table~\ref{Tab.DA_methods}, we compare the above DA methods by various aspects. 

\begin{itemize}
\item It is easy to find that nearly all paraphrasing-based and noising-based methods are not learnable, except for \textit{Seq2Seq} and \textit{Mixup}. However, most sampling-based methods are learnable except for the \textit{rule}-based ones. Learnable methods are usually more complex than non-learnable ones, thus sampling-based methods generate more diverse and fluent data than the former two.

\item Among all learnable methods, \textit{Mixup} is the only \textbf{online} one. That is to say, the DA process is during model training. Thus, \textit{Mixup} is the only one that outputs cross-label and discrete embedding from augmented data.

\item Comparing \textit{Learnable} and \textit{Resource}, we could see that most non-learnable methods require external knowledge resources which go beyond the original dataset and task definition. Commonly used resources include semantic thesauruses like WordNet and PPDB, handmade resources like misspelling dictionary in \cite{Coulombe2018TextDA[12]}, and artificial heuristics like the ones in \cite{Min2020SyntacticDA[13]} and \cite{Kang2018AdvEntuReAT[67]}.

\begin{table}[]\footnotesize
\caption{ Characteristics of different DA methods. \textit{Learnable} denotes whether the methods involve model training; \textit{online} and \textit{offline} denote whether the DA process is during or after model training. \textit{Ext.Know} denotes to whether the methods require external knowledge resources to generate augmented data. \textit{Pretrain} denotes whether the methods require a pre-trained model. \textit{Task-related} denotes whether the methods consider the label information, task format, and task requirements to generate augmented data. \textit{Level} denotes the depth and extent to which elements of the instance/data are modified by the DA; \textit{t}, \textit{e}, and \textit{l} denote text, embedding, and label, respectively. \textit{Granularity} indicates the extent to which the method could augment; \textit{w}, \textit{p}, and \textit{s} denote word, phrase, and sentence, respectively.}
\label{Tab.DA_methods}
\centering
\setlength\tabcolsep{2pt}
\begin{tabular}{cccccccc}
\hline
\hline
\multicolumn{2}{l}{}                      & Learnable   & Ext.Know  & Pretrain  & Task-related & Level & Granularity \\ \cline{4-4}
\hline
\multirow{6}{*}{\rotatebox{90}{Paraphrasing}} & Thesauruses   & \xmark   & \cmark       & \xmark      & \xmark   & $t$  & $w$    \\
                              & Semantic Embeddings    & \xmark  & \cmark      & \xmark    & \xmark   & $t$   & $w,p$    \\
                              & Language Models          & \xmark  & \xmark      & \cmark    & \xmark   & $t$   & $w$    \\
                              & Rules         & \xmark  & \cmark     & \xmark   & \xmark   & $t$   & $w,p,s$    \\
                              & Machine Translation            & \xmark      & \xmark     & \xmark & \xmark   & $t$   & $s$    \\
                              & Model Generation     & offline   & \xmark     & \xmark    & \cmark   & $t$    & $s$     \\
\hline
\multirow{5}{*}{\rotatebox{90}{Noising}}      & Swapping      & \xmark       & \xmark     & \xmark  & \xmark  & $t$    & $w,p,s$    \\
                              & Deletion      & \xmark       & \xmark      & \xmark   & \xmark  & $t$    & $w,p,s$   \\
                              & Insertion     & \xmark  & \cmark       & \xmark       & \xmark   & $t$   & $w,p,s$    \\
                              & Substitution  & \xmark  & \cmark      & \xmark  & \xmark   & $t$   & $w,p,s$   \\
\hline
\multirow{4}{*}{\rotatebox{90}{Sampling}}   & Rules         & \xmark   & \cmark         & \xmark     & \cmark   & $t,l$    & $w,p,s$    \\

                              & Non-pretrained     & offline   & \xmark         & \xmark       & \cmark   & $t,l$   & $s$    \\
                              & Pretrained    & offline   & \xmark         & \cmark      & \cmark   & $t,l$   & $s$     \\
                              & Self-training   & offline  & \xmark         & \xmark     & \cmark  & $t,l$   & $s$    \\
                              & Mixup        & online    & \xmark     &\xmark    & \cmark   & $e,l$    & $s$   \\

\hline
\hline
\end{tabular}
\end{table}

\item Through \textit{Learnable}, \textit{Ext.Know} and \textit{Pretrain}, it can be seen that in addition to artificial heuristics, DA requires other external interventions to generate valid new data. This includes model training objectives, external knowledge resources, and knowledge implicit in pretrained language models.

\item Comparing \textit{Learnable} and \textit{Task-related}, we could see that all paraphrasing-based and noising-based methods except model generation are not task-related. They generate augmented data given only original data without labels or task definition. However, all sampling-based methods are task-related because heuristics and model training are adopted to satisfy the needs of specific tasks.

\item  Comparing \textit{Level} and \textit{Task-related}, we could see that they are relevant. The paraphrasing-based methods are at the text level. The same is true for noising-based methods, except for Mixup, which augments both embeddings and labels. All sampling-based methods are at the text and label level since the labels are also considered and constructed during augmentation.

\item Comparing \textit{Learnable} and \textit{Granularity}, we could see that almost all non-learnable methods could be used for word-level and phrase-level DA, but all learnable methods could only be applied for sentence-level DA. Although learnable methods generate high-quality augmented sentences, unfortunately, they do not work for document augmentation because of their weaker processing ability for documents. Thus, document augmentation still relies on simple non-learnable methods, which is also a current situation we have observed in our research.
\end{itemize}

%Hi here!
%%\textbf{See the 2nd survey for level information!}

%\clearpage

\section{Strategies and Tricks}
\label{tricks}
The three types of DA methods including paraphrasing, noising, and sampling, as well as their characteristics, have been introduced above. In practical applications, the effect of the DA method is influenced by many factors. In this chapter, we present these factors to inspire our readers to use some strategies and tricks for selecting and constructing suitable DA methods. 
\subsection{Method Stacking}
The methods in Section~\ref{overview} are not mandatory to be applied alone. They could be combined for better performance. Common combinations include:

\paragraph{\textbf{The Same Type of Methods}}
Some works combine different paraphrasing-based methods and obtain different paraphrases, to increase the richness of augmented data. For example, Liu et al.~\cite{Liu2020DocumentlevelMS[7]} use both thesauruses and semantic embeddings, and Jiao et al.~\cite{Jiao2020TinyBERTDB[8]} use both semantic embeddings and MLMs. As for noising-based methods, the former unlearnable ways are usually used together like ~\cite{Peng2020DataAF[37]}. It is because these methods are simple, effective, and complementary. Some methods also adopt different sources of noising or paraphrasing like \cite{Regina[40]} and ~\cite{Xie2017DataNA[16]}. The combination of different resources could also improve the robustness of the model.

\paragraph{\textbf{Unsupervised Methods}}
In some scenarios, the simple and task-independent unsupervised DA methods could meet the demand. Naturally, they are grouped together and widely used. Wei et al.~\cite{Wei2019EDA[5]} introduce a DA toolkit called EDA that consists of synonym replacement, random insertion, random swap, and random deletion. EDA is very popular and used for many tasks (\cite{longpre-etal-2020-effective[34],Rastogi2020CanWA[39]}). UDA by Xie et al~\cite{xie2019unsupervised[9]} includes back-translation and unsupervised  noising-based methods; it is also used in many tasks like \cite{daval-frerot-weis-2020-wmd[33]}.

\paragraph{\textbf{Multi-granularity}}
Some works apply the same method at different levels to enrich the augmented data with changes of different granularities and improve the robustness of the model. For example, Wang et al.~\cite{wang2015s[6]} train both word embeddings and frame embeddings by Word2Vec; Guo et al.~\cite{guo2019augmenting[24]} apply Mixup at the word and sentence level, and Yu et al.~\cite{Yu2019HierarchicalDA[17]} use a series of noising-based methods at both the word and the sentence level.

\subsection{Optimization}
The optimization process of DA methods directly influences the quality of augmented data. We introduce it through four angles: the use of augmented data, hyperparameters, training strategies, and training objects.

\subsubsection{The Use of Augmented Data}
The way of using augmented data directly influences the final effect. From the perspective of data quality, the augmented data could be used to pre-train a model if it is not of high quality; otherwise, it could be used to train a model directly. From the perspective of data amount, if the amount of the augmented data is much higher than the original data, they are usually not directly used together for model training. Instead, some common practices include (1) oversampling the original data before training the model (2) pre-training the model with the augmented data and fine-tuning it on the original data.

\subsubsection{Hyperparameters}
All the above methods involve hyperparameters that largely affect the augmentation effect. We list some common hyperparameters in Figure~\ref{Fig.taxonomy_of_DA}:

\tikzstyle{leaf}=[%draw=hiddendraw,
    rounded corners,minimum height=1.2em,
    fill=myblue,text opacity=1,    align=center,
    fill opacity=.5,  text=black,align=left,font=\scriptsize,
inner xsep=3pt,
inner ysep=1pt,
]

\begin{figure*}[thp]
  \centering
\begin{forest}
  forked edges,
  for tree={
  grow=east,
  reversed=true,%increase counter-clockwise
  anchor=base west,
  parent anchor=east,
  child anchor=west,
  base=left,
  font=\scriptsize,
  rectangle,
  draw=black,%hiddendraw,
  rounded corners,align=left,
  minimum width=2.5em,
  minimum height=1.2em,
  %  edge+={darkgray, line width=1pt},
%  l sep+=2.5pt,
%  s sep+=-5pt,
s sep=6pt,
inner xsep=3pt,
inner ysep=1pt,
  },
  %before packing={where n children=3{calign child=2, calign=child edge}{}},
  %before typesetting nodes={where content={}{coordinate}{}},
  %where level<=1{line width=2pt}{line width=1pt},
  where level=1{text width=5em,font=\scriptsize}{},
  where level=2{text width=6.5em,font=\scriptsize}{},
  where level=3{text width=14em,font=\scriptsize}{},
  where level=4{font=\scriptsize}{},
  where level=5{font=\scriptsize}{},
  [Methods
    [Paraphrasing
        [1. Thesauruses\\2. Semantic\\~~~~Embeddings\\3. Language\\~~~~Models\\4. Rules
           [(1) Number of replacements\\(2) Probability of replacement
           ,leaf]
        ]
        [5. Machine\\~~~~Translation
           [(1) Number of (intermediate) languages\\(2) Types of (intermediate) languages
           ,leaf]
        ]
        [6. Model\\~~~~Generation
           [(1) Parameters in the neural network
           ,leaf]
        ]
    ]
    [Noising
        [1. Swapping\\2. Deletion\\3. Insertion\\4. Substitution
           [(1) Number of operations\\(2) Probability of operations
           ,leaf]
        ]
    ]
    [Sampling
        [1. Rules
           [(1) Number of replacements
           ,leaf]
        ]
        [2. Non\\~~~~-Pretrained\\3. Pretrained\\4. Self-training\\5. Mixup\\
           [(1) Parameters in the neural network
           ,leaf]
        ]
    ]
  ]
\end{forest}
\caption{Hyperparameters that affect the augmentation effect in each DA method.}
\label{Fig.taxonomy_of_DA}
\end{figure*}
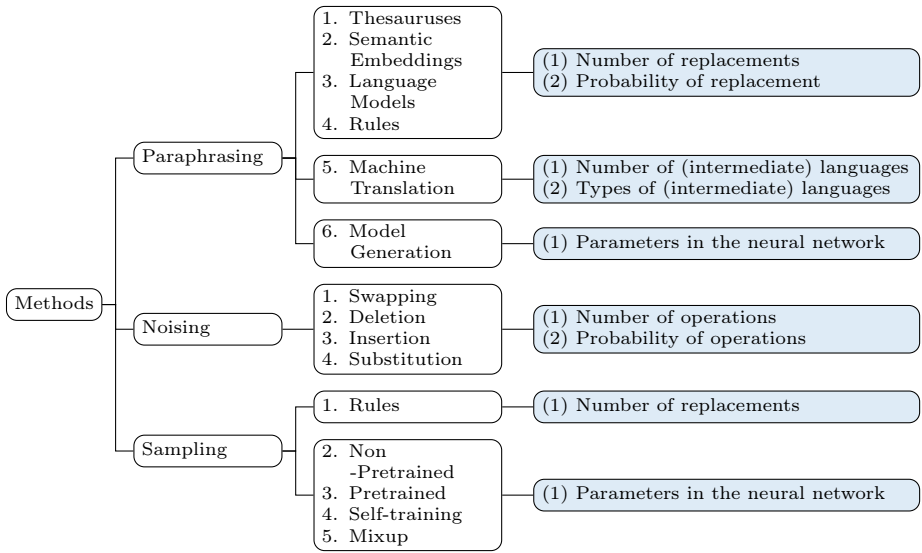

\subsubsection{Training Strategies}
Some works apply training strategies based on the basic data augmentation methods. For example, Qu et al.~\cite{Qu2021CoDACA[64]} combine back-translation with adversarial training. Similarly, Quteineh et al.~\cite{quteineh-etal-2020-textual[38]} transform the basic pre-trained model into an optimization problem~\footnote{Monte Carlo Tree Search.} to maximize the usefulness of the generated output. Hu et al.~\cite{Hu2019LearningDM[26]} and Liu et al.~\cite{Liu2020DataBT[81]} use pre-trained language models to generate augmented data, and transfer such progress into reinforcement learning. Some works (\cite{Rastogi2020CanWA[39], Shehnepoor2020GANgsterAF[58]}) take the idea of Generative Adversarial Networks to generate challenging augmented data.

\subsubsection{Training Objects}
Training objects are essential for model training, especially for the learnable DA methods. Nugent et al.~\cite{Nugent2020DetectingET[46]} propose a range of softmax temperature settings to ensure diversity while preserving semantic meaning. Hou et al.~\cite{Hou2020C2CGenDACG[19]} use duplication-aware attention and diverse-oriented regularization to generate more diverse sentences. Cheng et al.~\cite{cheng2020advaug[25]} employ curriculum learning to encourage the model to focus on the difficult training examples.

\subsection{Filtering}
Sometimes the progress of data augmentation inevitably introduces some noise even errors, thus filtering mechanisms are introduced to avoid this problem.

Some works filter input data in the initial stage to avoid inappropriate input affecting the augmentation effect. A typical example is sentence length, i.e., filter sentences that are too short (\cite{li2020conditional[22]}). Liu et al.~\cite{Liu2020ReverseOB[71]} filter out irrelevant numbers without augmenting them in solving Math Word Problems, to ensure the generated data is computationally right.

In addition, some works filter the synthetic augmented data at the end-stage. This is usually achieved through a model. For example, Zhang et al.~\cite{Zhang2020ParallelDA[11]} employ a discriminator to filter the back-translation results. Tavor et al.~\cite{anaby2019not[20]} and Peng et al.~\cite{Peng2020DataAF[37]} both apply a classifier to filter the augmented sentences generated by pre-trained models to ensure the data quality.

\section{Applications on NLP Tasks}
\label{applications}
Although a variety of data augmentation methods have emerged in the field of NLP in recent years, it is difficult to directly compare their performance. This is because different tasks, evaluation metrics, datasets, model architectures, and experimental settings make direct comparisons meaningless. Therefore, based on the work introduced above, we analyze the data augmentation methods from the perspective of different NLP tasks including text classification, text generation, and structured prediction~\cite{car-book}.

\begin{table}[]\footnotesize
\caption{The application of DA methods in NLP tasks. Note that if a paper involves multiple methods, we count it multiple times.}
\label{Tab.DA_NLP}
\centering
\begin{tabular}{ccp{2.5cm}<{\centering}p{2.3cm}<{\centering}p{2.3cm}<{\centering}}
\hline
\hline
\multicolumn{2}{l}{\multirow{2}{*}{}} & Text & Text & Structure   \\ % \cline{4-4}
\multicolumn{2}{l}{}                  &   Classification    &  Generation     &   Prediction                                  \\
\hline
\multirow{6}{*}{\rotatebox{90}{Paraphrasing}} 
								& Thesauruses  
									    %& \multicolumn{1}{m{3.5cm}}{ \cite{Zhang2015CharacterlevelCN[2]}, \cite{Wei2019EDAED[5]}, \cite{Liu2020DocumentlevelMS[7]}, \cite{Coulombe2018TextDA[12]}, \cite{daval-frerot-weis-2020-wmd[33]}, \cite{longpre-etal-2020-effective[34]}, \cite{Zhang2020OnDA[44]}, \cite{Zuo2020KnowDisKE[96]}, \cite{liu-yu-2020-blcu[150]}}
										& \cite{Zhang2015CharacterlevelCN[2]}, \cite{Wei2019EDA[5]}, \cite{Liu2020DocumentlevelMS[7]}, \cite{Coulombe2018TextDA[12]}, \cite{daval-frerot-weis-2020-wmd[33]}, \cite{longpre-etal-2020-effective[34]}, \cite{Zhang2020OnDA[44]}, \cite{Zuo2020KnowDisKE[96]}, \cite{liu-yu-2020-blcu[150]}, \cite{DBLP:conf/acl/KovatchevSLD20[259]}
										&  - 
										& \cite{daval-frerot-weis-2020-wmd[33]}, \cite{Dai2020AnAO[78]}  
										\\
                              & Embeddings   
										 &  \cite{wang2015s[6]}, \cite{Liu2020DocumentlevelMS[7]}, \cite{DBLP:conf/acl/KovatchevSLD20[259]}
										& - 
										& - 
										\\
                              & Language Models          
										& \cite{Regina[40]}, \cite{TapiaTllez2020DataAW[49]}, \cite{kobayashi-2018-contextual[174]}, \cite{Wu2019ConditionalBC[243]}, \cite{bari-etal-2021-uxla[252]}
										& \cite{fadaee-etal-2017-data[183]}
										&  - 
										\\
                              & Rules         
										& \cite{Regina[40]}, \cite{Coulombe2018TextDA[12]}, \cite{louvan-magnini-2020-simple[43]}
										& -
										& \cite{Sahin2018DataAV[169]}, \cite{andreas-2020-good[241]}
										\\
                              & Machine Translation            
										& \cite{daval-frerot-weis-2020-wmd[33]}, \cite{longpre-etal-2020-effective[34]}, \cite{Regina[40]}, \cite{xie2019unsupervised[9]}, \cite{Ibrahim2020AlexUBackTranslationTLAS[32]}, \cite{Rastogi2020CanWA[39]}, \cite{aroyehun-gelbukh-2018-aggression[165]}, \cite{Coulombe2018TextDA[12]}, \cite{Luque2019AtalayaAT[10]}, \cite{Barrire2020ImprovingSA[89]}, \cite{Lun2020MultipleDA[137]}, \cite{liu-yu-2020-blcu[150]}, \cite{risch-krestel-2018-aggression[167]}
										& \cite{Zhang2020ParallelDA[11]}, \cite{Fabbri2021ImprovingZA[77]}
										& \cite{daval-frerot-weis-2020-wmd[33]}, \cite{Yu2018QANetCL[249]}, \cite{Bornea2021MultilingualTL[48]}, \cite{longpre-etal-2019-exploration[201]}
										\\
                              & Model Generation     
										&  \cite{Liu2020TellMH[30]}, \cite{Kober2021DataAF[97]}, \cite{Xu2020DataAF[106]}, \cite{guo-etal-2020-text[153]}, \cite{zhao-etal-2019-data[217]}
										& \cite{Liu2020TellMH[30]}, \cite{Wan2020ImprovingGE[105]}, \cite{kumar-etal-2019-submodular[221]}, \cite{Li2019InsufficientDC[242]}
										& \cite{Liu2020TellMH[30]}, \cite{Hou2018SequencetoSequenceDA[18]}, \cite{Hou2020C2CGenDACG[19]}, \cite{li2020conditional[22]}, \cite{Yoo2020VariationalHD[87]}, \cite{Yin2020DialogST[143]}   
										\\
\hline
\multirow{5}{*}{\rotatebox{90}{Noising}}      
								& Swapping      
										&  \cite{Wei2019EDA[5]}, \cite{longpre-etal-2020-effective[34]}, \cite{Zhang2020OnDA[44]}, \cite{Rastogi2020CanWA[39]}, \cite{Yan2019DataAF[15]}, \cite{Luque2019AtalayaAT[10]}, \cite{du-black-2018-data[168]}, \cite{DBLP:conf/acl/KovatchevSLD20[259]}
										& -
										& \cite{Dai2020AnAO[78]}
										\\
                              & Deletion      
										& \cite{Wei2019EDA[5]}, \cite{longpre-etal-2020-effective[34]}, \cite{Zhang2020OnDA[44]}, \cite{Rastogi2020CanWA[39]}, \cite{Yan2019DataAF[15]}, \cite{Yu2019HierarchicalDA[17]}, \cite{chen-etal-2021-hiddencut[255]}
										& \cite{Peng2020DataAF[37]}
										& -
										\\ 
                              & Insertion     
										& \cite{Wei2019EDA[5]}, \cite{longpre-etal-2020-effective[34]}, \cite{Zhang2020OnDA[44]}, \cite{Rastogi2020CanWA[39]}, \cite{DBLP:conf/acl/KovatchevSLD20[259]}
										& \cite{Peng2020DataAF[37]}
										& - 
										\\
                              & Substitution  
										&  \cite{daval-frerot-weis-2020-wmd[33]}, \cite{Regina[40]}, \cite{xie2019unsupervised[9]}, \cite{Coulombe2018TextDA[12]}, \cite{Lun2020MultipleDA[137]} 
										& \cite{Xie2017DataNA[16]}, \cite{wang-etal-2018-switchout[170]}, \cite{Peng2020DataAF[37]}
										& \cite{daval-frerot-weis-2020-wmd[33]}, \cite{louvan-magnini-2020-simple[43]}, \cite{Dai2020AnAO[78]}, \cite{Shi2021SubstructureSS[117]}
										\\
\hline
\multirow{4}{*}{\rotatebox{90}{Sampling}}    
								& Rules         
										&  \cite{Min2020SyntacticDA[13]}, \cite{Kang2018AdvEntuReAT[67]}, \cite{Kober2021DataAF[97]}, \cite{Shakeel2020AMM[128]}, \cite{Lun2020MultipleDA[137]}, \cite{Xu2016ImprovedRC[191]}, \cite{chen-etal-2021-cross-language[254]}, \cite{jiang-etal-2021-cori[257]} 
										& \cite{Mou2020MultimodalDS[42]}, \cite{Zhang2020DialogueDO[85]}, \cite{Asai2020LogicGuidedDA[144]}, \cite{bergmanis-goldwater-2019-data[219]}
										& \cite{Zmigrod2019CounterfactualDA[204]}
										\\ 
                              & Non-Pretrained    
										& \cite{Kang2018AdvEntuReAT[67]}, \cite{Raille2020FastCD[54]}, \cite{Yoo2019DataAF[161]}, \cite{Zhou2020ForecastingET[130]}, \cite{niu-bansal-2019-automatically[216]}
										& \cite{Zhang2020ParallelDA[11]}, \cite{Chen2020PatternawareDA[65]}, \cite{Yao2020DomainTB[107]}, \cite{ijcai2020-496[152]}, \cite{Sennrich2016ImprovingNM[244]}
										& \cite{Yoo2019DataAF[161]}, \cite{liu-etal-2021-mulda[260]}
										\\
                              & Pretrained    
										& \cite{Zhang2020OnDA[44]}, \cite{Ng2020SSMBASM[36]}, \cite{kumar2020data[21]}, \cite{quteineh-etal-2020-textual[38]}, \cite{Liu2020DataBT[81]}, \cite{anaby2019not[20]}, \cite{Abonizio2020PretrainedDA[93]}, \cite{Staliunaite2021ImprovingCC[118]}, \cite{dong-etal-2021-data[251]}
										& \cite{Peng2020DataAF[37]}, \cite{Ng2020SSMBASM[36]}, \cite{tarjan2020deep[41]}
										& \cite{Peng2020DataAF[37]}, \cite{Riabi2020SyntheticDA[62]} 
										\\
                              & Self-training   
										&  \cite{Du2021SelftrainingIP[69]}, \cite{Yang2020NeuralRF[59]}, \cite{Aleksandr-inproceedings[50]}, \cite{Miao2020TwitterDA[35]}
										& \cite{Montella2020DenoisingPA[47]}, \cite{xu-etal-2021-augnlg[263]}
										& \cite{Yang2020NeuralRF[59]}
										\\
							& Mixup        
										& \cite{guo2019augmenting[24]}, \cite{Sun2020MixupTransformerDD[75]}, \cite{Si2020BetterRB[83]}, \cite{si-etal-2021-better[262]}
										& \cite{cheng2020advaug[25]}
										& \cite{Chen2020LocalAB[72]}
										\\

\hline
\hline
\end{tabular}
\end{table}

\begin{itemize}
\item Text classification is the simplest and most basic natural language processing problem. That is, for a piece of text input, output the category to which the text belongs, where the category is a pre-defined closed set.\footnote{Text matching tasks such as Natural Language Inference can also be transformed into text classification.}

\item Text generation, as the name implies, is to generate the corresponding text given the input data. The most classic example is machine translation.

\item The structured prediction problem is usually unique to NLP. Different from the text classification, there are strong correlation and format requirements between the output categories in the structured prediction problem.

\end{itemize}
In this section, we try to analyze the features as well as the development status of DA in these tasks. Some statistical results are shown in Table~\ref{Tab.DA_NLP} and Table~\ref{tab:timeline}.

DA methods are applied more widely in text classification than other NLP tasks in general and in each category. Moreover, each individual DA method could be applied to text classification. Such application advantage is because of the simple form of text classification: given the input text, it directly investigates the model's understanding of semantics by label prediction. Therefore, it is relatively simple for data augmentation to only consider retaining the semantics of words that are important for classification.

As for text generation, it prefers sampling-based methods to bring more semantic diversity. And structured prediction prefers paraphrasing-based methods because it is sensitive to data format. Thus, it has higher requirements for data validity.

By comparing each DA method, we can see that simple and effective unsupervised methods, including machine translation, thesaurus-based paraphrasing, and random substitution, are quite popular. In addition, learnable methods like paraphrasing-based model generation and  sampling-based pretrained models, also gain a lot of attention because of their diversity and effectiveness.

We also show the development process of the DA method on three types of tasks through a timeline (Table~\ref{tab:timeline}). On the whole, the number of applications of DA in these tasks has increased these years. Text classification is the first task to use DA, and the number of corresponding papers is also larger than the other two tasks. In terms of text generation and structured prediction, DA is receiving increasing attention. Paraphrasing-based methods have always been a popular method. In recent years, sampling-based methods show clear momentum in text classification and text generation, because they bring more gains to powerful pretrained language models than paraphrasing-based methods. However, people still tend to use paraphrasing and noising-based methods in structured prediction.
%%%
%According to the difference of input data, text generation tasks can be roughly divided into the following three categories: 1) text-to-text generation; 2) data-to-text generation; 3) image-to-text generation.

%\iffalse
\begin{table}\tiny
\centering
%\begin{center}
%\begin{longtable}

%大事表居中显示
\renewcommand\arraystretch{1.4}\arrayrulecolor{baseD}
\captionsetup{singlelinecheck=false, labelfont=sc, labelsep=quad}
\caption{Timeline of DA methods applied in three kinds of NLP tasks. The time for each paper is based on its first arXiv version (if exists) or estimated submission time. \textcolor{red}{P} denotes paraphrasing-based methods; \textcolor{blue}{N} denotes noising-based methods; \textcolor{orange}{S} denotes sampling-based methods.}\vskip -1.5ex
%vskip是标题到下面的表格的距离
%\begin{tabular}{@{\,}r <{\hskip 2pt} !{\foo} >{\raggedright\arraybackslash}p{5cm}}
%\begin{tabular}{lp{3cm}p{2.5cm}p{2.5cm}}
\textbf{~~~~~~~~Text Classification~~~~~~~~~~~~~~~~Text Generation~~~~~~~~~~~~~~~~~Structured Prediction}\\
%\end{tabular}
\begin{tabular}{@{\,}r <{\hskip 2pt} !{\foo}>{\raggedright\arraybackslash}p{3cm}@{\,}r <{\hskip 2pt} !{\foo} >{\raggedright\arraybackslash}p{2.5cm}@{\,}r <{\hskip 2pt} !{\foo} >{\raggedright\arraybackslash}p{2.5cm}}

%5cm这个参数是指表的总的宽度，如果想做得大一点就把5cm调整放大一点。
\toprule
\addlinespace[1.5ex]
%这个是表格的行距

2015.09 & Zhang et al.~\citep{Zhang2015CharacterlevelCN[2]} \textcolor{red}{P}  &   &  & & \\
		 & Wang et al.~\citep{wang2015s[6]} \textcolor{red}{P}  &   &  & & \\
2015.11 &  &   & Sennrich et al.~\cite{Sennrich2016ImprovingNM[244]} \textcolor{orange}{S}  & & \\
2016.01 & Xu et al.~\cite{Xu2016ImprovedRC[191]} \textcolor{orange}{S} & & & & \\
...  & &  & & \\
2017.03 &  &   & Xie et al.~\citep{Xie2017DataNA[16]} \textcolor{blue}{N}  & & \\
2017.05 &  &   & Fadaee et al.~\citep{fadaee-etal-2017-data[183]} \textcolor{red}{P}  & & \\
...  & &  & & \\
2018.04 &  &   & &  & Yu et al.~\citep{Yu2018QANetCL[249]} \textcolor{red}{P} \\
2018.05 & Kang et al.~\cite{Kang2018AdvEntuReAT[67]} \textcolor{orange}{S}&   & &  & \\
2018.06 & Kobayashi et al.~\cite{kobayashi-2018-contextual[174]} \textcolor{red}{P} & & & & \\
2018.07 &  &   & &  & Hou et al.~\citep{Hou2018SequencetoSequenceDA[18]} \textcolor{red}{P} \\
2018.08 & Aroyehun et al.~\cite{aroyehun-gelbukh-2018-aggression[165]}  \textcolor{red}{P} &   & Wang et al.~\citep{wang-etal-2018-switchout[170]} \textcolor{blue}{N}  & & \\
 & Risch et al.\cite{risch-krestel-2018-aggression[167]} \textcolor{red}{P} &   &  & & \\

2018.09 & Yoo et al.~\cite{Yoo2019DataAF[161]} \textcolor{orange}{S} & & & &  Yoo et al.~\cite{Yoo2019DataAF[161]} \textcolor{orange}{S} \\
2018.10 & Du et al.~\cite{du-black-2018-data[168]} \textcolor{blue}{N} & & & & Sahin et al.~\cite{Sahin2018DataAV[169]} \textcolor{red}{P}\\
2018.12 & Coulombe et al.~\citep{Coulombe2018TextDA[12]} \textcolor{red}{P}, \textcolor{blue}{N}  &   &  & & \\
 & Wu et al.~\cite{Wu2019ConditionalBC[243]} \textcolor{red}{P}  &   &  & & \\
2019.01 & Wei et al.~\citep{Wei2019EDA[5]} \textcolor{red}{P}, \textcolor{blue}{N}  &   &  & & \\
2019.04 & Xie et al.~\citep{xie2019unsupervised[9]} \textcolor{red}{P}, \textcolor{blue}{N}  &   &  & & \\
2019.05 & Guo et al.~\citep{guo2019augmenting[24]} \textcolor{orange}{S}  &   & Gao et al.~\cite{gao-etal-2019-soft[202]} \textcolor{blue}{N}  & & \\
2019.06 & & & Xia et al.~\cite{xia-etal-2019-generalized[203]} \textcolor{orange}{S} & & \\
 & & & Bergmanis et al.~\cite{bergmanis-goldwater-2019-data[219]} \textcolor{orange}{S} & & \\
 & & & Kumar et al.~\cite{kumar-etal-2019-submodular[221]} \textcolor{red}{P} & & \\
2019.07 & Yu et al.~\citep{Yu2019HierarchicalDA[17]} \textcolor{blue}{N}  &   & Li et al.~\cite{Li2019InsufficientDC[242]}  \textcolor{red}{P} & & Zmigrod et al.~\cite{Zmigrod2019CounterfactualDA[204]} \textcolor{orange}{S} \\
2019.08 & & & & & Yin et al.~\cite{Yin2020DialogST[143]} \textcolor{red}{P} \\
2019.09 & Luque et al.~\citep{Luque2019AtalayaAT[10]} \textcolor{red}{P}, \textcolor{blue}{N}  &   &  & & \\
		 & Yan et al.~\citep{Yan2019DataAF[15]} \textcolor{blue}{N}  &   &  & & \\ 

\midrule
\end{tabular}\label{tab:timeline}
\end{table}
%\end{longtable}
%\end{center}
%\fi

\clearpage

\begin{table}\tiny
\centering

\renewcommand\arraystretch{1.4}\arrayrulecolor{baseD}
\captionsetup{singlelinecheck=false, labelfont=sc, labelsep=quad}
\vskip -1.5ex
%vskip是标题到下面的表格的距离
%\begin{tabular}{@{\,}r <{\hskip 2pt} !{\foo} >{\raggedright\arraybackslash}p{5cm}}
%\begin{tabular}{lp{3cm}p{2.5cm}p{2.5cm}}
\textbf{~~~~~~~~Text Classification~~~~~~~~~~~~~~~~Text Generation~~~~~~~~~~~~~~~~~Structured Prediction}\\
%\end{tabular}
\begin{tabular}{@{\,}r <{\hskip 2pt} !{\foo}>{\raggedright\arraybackslash}p{3cm}@{\,}r <{\hskip 2pt} !{\foo} >{\raggedright\arraybackslash}p{2.5cm}@{\,}r <{\hskip 2pt} !{\foo} >{\raggedright\arraybackslash}p{2.5cm}}

%5cm这个参数是指表的总的宽度，如果想做得大一点就把5cm调整放大一点。
\toprule
\addlinespace[1.5ex]
%这个是表格的行距
2019.11 & Anaby et al.~\citep{anaby2019not[20]} \textcolor{orange}{S}  &   &  & & Longpre et al.~\cite{longpre-etal-2019-exploration[201]} \textcolor{red}{P} \\
 & Malandrakis et al.~\cite{malandrakis-etal-2019-controlled[214]} \textcolor{red}{P} & & & &  \\
 & Niu et al.~\cite{niu-bansal-2019-automatically[216]} \textcolor{orange}{S} & & & &  \\
 & Zhao et al.~\cite{zhao-etal-2019-data[217]} \textcolor{red}{P} & & & &  \\
2019.12 & Shakeel et al.~\cite{Shakeel2020AMM[128]} \textcolor{orange}{S} &  & & & \\
2020.01 &  &   &  & & Yoo et al.~\cite{Yoo2020VariationalHD[87]} \textcolor{red}{P}\\
2020.03 & Kumar et al.~\citep{kumar2020data[21]} \textcolor{orange}{S}  &   &  & & \\
		& Raille et al.~\citep{Raille2020FastCD[54]} \textcolor{orange}{S}  &   &  & & \\
2020.04 & Lun et al.~\cite{Lun2020MultipleDA[137]} \textcolor{red}{P}, \textcolor{blue}{N}, \textcolor{orange}{S} &   &Peng et al.~\cite{Peng2020DataAF[37]} \textcolor{blue}{N}, \textcolor{orange}{S} & & Li et al.~\citep{li2020conditional[22]} \textcolor{red}{P}   \\
		& &   &  & & Peng et al.~\cite{Peng2020DataAF[37]} \textcolor{orange}{S}   \\
2020.05 & Kober et al.~\cite{Kober2021DataAF[97]} \textcolor{red}{P}, \textcolor{orange}{S} &   & Zhang et al.~\citep{Zhang2020ParallelDA[11]} \textcolor{red}{P}, \textcolor{orange}{S}  & & \\
		& Cao et al.~\cite{Cao2020HateGANAG[142]} \textcolor{orange}{S} & & & & \\
2020.06 & Liu et al.~\cite{Liu2020DocumentlevelMS[7]} \textcolor{red}{P} &   &   Cheng et al.~\citep{cheng2020advaug[25]} \textcolor{orange}{S} & & \\
		& Qin et al.~\cite{Qin2020CoSDAMLMC[103]}  \textcolor{blue}{N} & & & & Qin et al.~\cite{Qin2020CoSDAMLMC[103]}  \textcolor{blue}{N} \\
2020.07 & Min et al.~\cite{Min2020SyntacticDA[13]} \textcolor{orange}{S}  &   &   Chen et al.~\citep{ijcai2020-496[152]} \textcolor{orange}{S}& & Andreas et al.~\cite{andreas-2020-good[241]} \textcolor{red}{P} \\
		&  Rastogi et al.~\cite{Rastogi2020CanWA[39]} \textcolor{red}{P}, \textcolor{blue}{N}  &   & Tarjan et al.~\cite{tarjan2020deep[41]} \textcolor{orange}{S} & & \\
		&  Regina et al.~\cite{Regina[40]} \textcolor{red}{P}, \textcolor{blue}{N}  &   & Mou et al.~\cite{Mou2020MultimodalDS[42]} \textcolor{orange}{S} & & \\

		& Asai et al.~\cite{Asai2020LogicGuidedDA[144]} \textcolor{orange}{S} & & & & \\
2020.09 & Ng et al.~\citep{Ng2020SSMBASM[36]} \textcolor{orange}{S}  &   &   Ng et al.~\citep{Ng2020SSMBASM[36]} \textcolor{orange}{S} & & Yang et al.~\cite{Yang2020NeuralRF[59]}  \textcolor{orange}{S} \\
		 & Zhang et al.~\citep{Zhang2020OnDA[44]} \textcolor{red}{P},\textcolor{blue}{N}, \textcolor{orange}{S}  &   &Zhang et al.~\cite{Zhang2020DialogueDO[85]} \textcolor{orange}{S} & &  \\
		 
2020.10 & Barrire et al.~\citep{Barrire2020ImprovingSA[89]} \textcolor{red}{P}  &   & Fabbri et al.~\cite{Fabbri2021ImprovingZA[77]} \textcolor{red}{P} &  & Liu et al.~\cite{Liu2020TellMH[30]} \textcolor{red}{P} \\
		& Louvan et al.~\citep{louvan-magnini-2020-simple[43]} \textcolor{red}{P}  &   &   &  & Louvan et al.~\citep{louvan-magnini-2020-simple[43]} \textcolor{blue}{N} \\
		& Tapia-T{\'e}llez et al.~\cite{TapiaTllez2020DataAW[49]}  \textcolor{red}{P} &  & & & Chen et al.~\cite{Chen2020LocalAB[72]} \textcolor{orange}{S} \\
		& Sun et al.~\cite{Sun2020MixupTransformerDD[75]} \textcolor{orange}{S} &  & & &Dai et al.~\cite{Dai2020AnAO[78]} \textcolor{red}{P}, \textcolor{blue}{N}\\
		& Abonizio et al.~\cite{Abonizio2020PretrainedDA[93]}  \textcolor{orange}{S} &  &  &  & Riabi et al.~\cite{Riabi2020SyntheticDA[62]} \textcolor{orange}{S}\\
		& Zuo et al.~\cite{Zuo2020KnowDisKE[96]}  \textcolor{red}{P} &  & &  & \\
2020.11 & Longpre et al.~\citep{longpre-etal-2020-effective[34]} \textcolor{red}{P}, \textcolor{blue}{N}  &   &   &  &  \\
		& Quteineh et al.~\citep{quteineh-etal-2020-textual[38]} \textcolor{orange}{S}  &   &   &  &  \\
2020.12 &  Miao et al.~\citep{Miao2020TwitterDA[35]} \textcolor{orange}{S}  &   & Wan et al.~\cite{Wan2020ImprovingGE[105]} \textcolor{red}{P} &  & Bornea et al.~\cite{Bornea2021MultilingualTL[48]} \textcolor{red}{P} \\
		& Daval et al.~\cite{daval-frerot-weis-2020-wmd[33]} \textcolor{red}{P} ,\textcolor{blue}{N}  &   & Yao et al.~\cite{Yao2020DomainTB[107]}  &  & Hou et al.~\citep{Hou2020C2CGenDACG[19]} \textcolor{red}{P} \\
		& Liu et al.~\cite{Liu2020DataBT[81]} \textcolor{orange}{S} &   & Montella et al.~\cite{Montella2020DenoisingPA[47]} \textcolor{orange}{S} \textcolor{orange}{S} &  & Daval et al.~\cite{daval-frerot-weis-2020-wmd[33]} \textcolor{red}{P} ,\textcolor{blue}{N} \\
		& Aleksandr et al.~\cite{Aleksandr-inproceedings[50]}  \textcolor{orange}{S} &  & Chen et al.~\cite{Chen2020PatternawareDA[65]} \textcolor{orange}{S}&  & \\
		& Si et al.~\cite{Si2020BetterRB[83]} \textcolor{orange}{S} & & & & \\
		& Xu et al.~\cite{Xu2020DataAF[106]}  \textcolor{red}{P} & & & & \\
		& Liu et al.~\cite{liu-yu-2020-blcu[150]} \textcolor{red}{P} & & & & \\
		& Guo et al.~\cite{guo-etal-2020-text[153]} \textcolor{red}{P} & & & & \\
		& Si et al.~\cite{si-etal-2021-better[262]} \textcolor{orange}{S} & & & & \\
		
		2021.01 & Shi et al.~\cite{Shi2021SubstructureSS[117]} \textcolor{blue}{N} &  & & & Shi et al.~\cite{Shi2021SubstructureSS[117]} \textcolor{blue}{N} \\
		 & Staliunaite et al.~\cite{Staliunaite2021ImprovingCC[118]} \textcolor{orange}{S} & & & & \\
		 & Dong et al.~\cite{dong-etal-2021-data[251]} \textcolor{orange}{S} &  & & &  \\
		2021.06 & Chen et al.~\cite{chen-etal-2021-cross-language[254]} \textcolor{orange}{S} &  & Xu et al.~\cite{xu-etal-2021-augnlg[263]}  \textcolor{orange}{S} & &  \\
		 & Chen et al.~\cite{chen-etal-2021-hiddencut[255]} \textcolor{blue}{N} &  & & &  \\
		 & Jiang et al.~\cite{jiang-etal-2021-cori[257]} \textcolor{orange}{S} &  & & &  \\
		 & Kovatchev et al.~\cite{DBLP:conf/acl/KovatchevSLD20[259]} \textcolor{red}{P}, \textcolor{blue}{N} &  & & &  \\
		2021.08 & Bari et al.~\cite{bari-etal-2021-uxla[252]} \textcolor{red}{P} &  & & & Liu et al.~\cite{liu-etal-2021-mulda[260]} \textcolor{orange}{S}\\

\bottomrule
\end{tabular}
\end{table}

\clearpage

\section{Related Topics}
\label{topics}
How does data augmentation relate to other learning methods? In this section, we connect data augmentation with other similar topics.

\subsection{Pretrained Language Models}
The training of most pre-trained language models (PLMs) is based on self-supervised learning. Self-supervised learning mainly uses auxiliary tasks to mine its supervised information from large-scale unsupervised data, and trains the network through this constructed supervised information, so that it can learn valuable representations for downstream tasks. From this perspective, PLMs also introduce more training data into downstream tasks, in an implicit way. On the other hand, the general large-scale unsupervised data of PLMs may be out-of-domain for specific tasks. Differently, the task-related data augmentation methods essentially focus on specific tasks.

\subsection{Contrastive Learning}
Contrastive learning is to learn an embedding space in which similar samples are close to each other while dissimilar ones are far apart. It focuses on learning the common features between similar samples and distinguishing the differences between dissimilar ones. The first step of contrastive learning is applying data augmentation to construct similar samples with the same label, and the second step is to randomly choose instances as the negative samples. Thus, contrastive learning is one of the applications of data augmentation.

\subsection{Other Data Manipulation Methods}
In addition to DA, there are some other data manipulation methods to improve model generalization~\cite{kukavcka2017regularization,Hu2019LearningDM[26]}. \textit{Oversampling} is usually used in data imbalance scenarios. It simply samples original data from the minority group as new samples, instead of generating augmented data. \textit{Data cleaning} is additionally applied to the original data to improve data quality and reduce data noise. It usually includes lowercasing, stemming, lemmatization, etc. \textit{Data weighting} assigns different weights to different samples according to their importance during training, without generating new data. \textit{Data synthesis} provides entire labeled artificial examples instead of augmented data generated by models or rules.

\subsection{Generative Adversarial Networks}
Generative Adversarial Networks (GANs) are first introduced by Goodfellow et al.~\cite{Goodfellow2014GenerativeAN}. As a type of semi-supervised method, GANs include the generative model, which is mainly used to challenge the discriminator of GANs, while the generative models in some DA methods are directly used to augment training data. Moreover, the generative model of GANS is applied as a DA method in some scenes like \cite{Rastogi2020CanWA[39],morris-etal-2020-textattack[139],Shehnepoor2020GANgsterAF[58],Kober2021DataAF[97],Zhou2020ForecastingET[130],Cao2020HateGANAG[142]}, and have demonstrated to be effective for data augmentation purposes. 
\subsection{Adversarial Attacks}
Adversarial attacks are techniques to generate adversarial examples attacking a machine learning model, i.e., causing the model to make a mistake. Some works use DA methods like code-switch substitution to generate adversarial examples as consistency regularization~\cite{Zheng2021ConsistencyRF[248]}.

\section{Challenges and Opportunities}
\label{challenges}
Data augmentation has seen a great process over the last few years, and it has provided a great contribution to large-scale model training as well as the development of downstream tasks. Despite the process, there are still challenges to be addressed. In this section, we discuss some of these challenges and future directions that could help advance the field.

\paragraph{\textbf{Theoretical Narrative}} At this stage, there appears to be a lack of systematic probing work and theoretical analysis of DA methods in NLP. The few related works are of DA in the image domain, considering data augmentation as encoding a priori knowledge about data or task invariance~\cite{dao2019kernel}, variance reduction~\cite{chen2020group} or regularization methods~\cite{wu2020generalization}.  In NLP, Most previous works propose new methods or prove the effectiveness of the DA method on downstream tasks, but do not explore the reasons and laws behind it, e.g., from the perspective of mathematics. The discrete nature of natural language makes theoretical narrative essential since narrative helps us understand the nature of DA, without being limited to determining effectiveness through experiments.

\paragraph{\textbf{More Exploration on Pretrained Language Models}} In recent years, pre-trained language models have been widely applied in NLP, which contain rich knowledge through self-supervision on a huge scale of corpora. There are works using pre-trained language models for DA, but most of them are limited to [MASK] completion~\cite{Wu2019ConditionalBC[243]}, direct generation after fine-tuning~\cite{Zhang2020OnDA[44]}, or self-training~\cite{Du2021SelftrainingIP[69]}. Is DA still helpful in the era of pre-trained language models? Or, how to further use the information in pre-trained models to generate more diverse and high-quality data with less cost? There are some initial explorations in these directions~\cite{zhou-etal-2022-flipda,liu2022wanli}, while we still look forward to more works in the future.

\paragraph{\textbf{Few-shot Scenarios}} In few-shot scenarios, models are required to achieve performance which rivals that of traditional machine learning models, yet the amount of training data is extremely limited. DA methods provide a direct solution to the problem. However, most current works in few-shot scenarios are paraphrasing-based methods~\cite{Fabbri2021ImprovingZA[77]}. Such methods ensure the validity of the augmented data, but also lead to insufficient semantic diversity. Mainstream pretrained language models obtain rich semantic knowledge by language modeling. Such knowledge even covers to some extent the semantic information introduced by traditional paraphrasing-based DA methods. In other words, the improvement space that traditional DA methods bring to pretrained language models has been greatly compressed. Therefore, it is an interesting question how to provide models with fast generalization and problem solving capability by generating high quality augmented data in few-shot scenarios.

\paragraph{\textbf{Retrieval Augmentation}} Retrieval-augmented language models integrate retrieval into pre-training and downstream usage~\cite{singh2021end,yogatama2021adaptive}.\footnote{See \url{https://ruder.io/ml-highlights-2021/} for further information.}
Retrieval augmentation makes models much more parameter-efficient, as they need to store less knowledge in their parameters and can instead retrieve it. It also enables efficient domain adaptation by simply updating the data used for retrieval~\cite{khandelwal2019generalization}.
Recently, the size of the retrieval corpora has achieved explosive growth~\cite{borgeaud2021improving} and models have been equipped with the ability to query the web for answering questions~\cite{komeili2021internet,nakano2021webgpt}.
In the future, there may be different forms of retrieval to leverage different kinds of information such as common sense knowledge, factual relations, linguistic information, etc. Retrieval augmentation could also be combined with more structured forms of knowledge retrieval, such as methods from knowledge base population and open information extraction.

\paragraph{\textbf{More Generalized Methods for NLP}} Natural language is most different from image or sound in that its representation is discrete. At the same time, NLP includes specific tasks such as structured prediction that are not available in other modalities. Therefore, unlike general methods such as \textit{clipping} for image augmentation or \textit{speed perturbation} for audio augmentation, there is currently no DA method that can be effective for all NLP tasks. This means that there is still a gap for DA methods between different NLP tasks. With the development of pre-trained models, this seems to have some possibilities. Especially the proposal of T5~\cite{Raffel2020ExploringTL} and GPT3~\cite{Brown2020LanguageMA}, as well as the emergence of prompting learning further verify that the formalization of tasks in natural language can be independent of the traditional categories, and a more generalized model could be obtained by unifying task definitions. 

\paragraph{\textbf{Working with Long Texts and Low Resources Languages}} The existing methods have made significant progress in short texts and common languages. However, limited by model capabilities, DA methods on long texts still struggle with the simplest methods of paraphrasing and noising~\cite{Liu2020DocumentlevelMS[7],Yan2019DataAF[15],Yu2019HierarchicalDA[17]} (as shown in Table~\ref{Tab.DA_methods}). At the same time, limited by data resources, augmentation methods of low resource languages are scarce~\cite{kumar2020data[21]}, although they have more demand for data augmentation. Obviously, exploration in these two directions is still limited, and they could be promising directions.

\section{Conclusion}
\label{conclusion}
In this paper, we presented a comprehensive and structured survey of data augmentation for natural language processing. In order to inspect the nature of DA, we framed DA methods into three categories according to the \textbf{diversity} of augmented data, including paraphrasing, noising, and sampling. Such categories help to understand and develop DA methods. We also introduced the characteristics of DA methods and their applications in NLP tasks, then analyzed them through a timeline. In addition, we introduced some tricks and strategies so that researchers and practitioners can refer to obtain better model performance. Finally, we distinguish DA with some related topics and outlined current challenges as well as opportunities for future research. 

% \section*{References}

\bibliography{9_mybibfile}

\newpage
\appendix
\section{Related Resources}
\label{appendix}
\includegraphics[scale=0.02]{icon_paraphrasing.png}
\includegraphics[scale=0.02]{icon_noising.png}
\includegraphics[scale=0.02]{icon_sampling.png} 
There are some \textbf{popular resources} that provides helpful information or APIs of DA.
\begin{itemize}
    \item[-] A Visual Survey of Data Augmentation in NLP (\href{https://amitness.com/2020/05/data-augmentation-for-nlp/}{Blog})
    \item[-] EDA: Easy Data Augmentation Techniques for Boosting Performance on Text Classification Tasks (\href{https://github.com/jasonwei20/eda_nlp}{Repo})
    \item[-] Unsupervised Data Augmentation
    (\href{https://github.com/google-research/uda}{Repo})
    \item[-] Unsupervised Data Augmentation (Pytorch)
    (\href{https://github.com/SanghunYun/UDA_pytorch}{Repo})
    \item[-] nlpaug: Data Augmentation in NLP
    (\href{https://github.com/makcedward/nlpaug}{Repo})
    \item[-] TextAttack: Generating Adversarial Examples for NLP Models
    (\href{https://github.com/QData/TextAttack}{Repo})
    \item[-] AugLy: A Data Augmentations Library for Audio, Image, Text, and Video
    (\href{https://github.com/facebookresearch/AugLy}{Repo})
    \item[-] NL-Augmenter: A Collaborative Repository of Natural Language Transformations
    (\href{https://github.com/GEM-benchmark/NL-Augmenter/}{Repo})
\end{itemize}

\includegraphics[scale=0.02]{icon_paraphrasing.png}
\includegraphics[scale=0.02]{icon_noising.png}
\includegraphics[scale=0.02]{icon_sampling.png} 
In addition to English, there are resources in \textbf{other languages} such as:
\begin{itemize}
    \item[-] Turkish: nlpaug: Data Augmentation in NLP (\href{https://github.com/makcedward/nlpaug}{Repo})
    \item[-] Chinese: NLP Data Augmentation with EDA, BERT, BART, and back-translation (\href{https://github.com/InsaneLife/NLPDataAugmentation}{Repo})
    \item[-] Chinese: ChineseNLPDataAugmentation4Paddle: NLP Data Augmentation with EDA and BERT Contextual Augmentation, Customized for PaddleNLP (\href{https://github.com/RicardoL1u/ChineseNLPDataAugmentation4Paddle}{Repo})
    \item[-] Chinese: Tencent AI Lab Embedding Corpus for Chinese Words and Phrases (\href{https://ai.tencent.com/ailab/nlp/en/embedding.html}{Link})
    
\end{itemize}

\end{document}